\documentclass{article}
\ifdefined\pdfminorversion
\fi
\usepackage[utf8]{inputenc}
\usepackage{main}
\usepackage{microtype}
\usepackage{times}

\usepackage{amsmath,amsfonts,bm}

\renewcommand{\eqref}[1]{(\ref{#1})}

\def\1{\bm{1}}

\DeclareMathAlphabet{\mathsfit}{\encodingdefault}{\sfdefault}{m}{sl}
\SetMathAlphabet{\mathsfit}{bold}{\encodingdefault}{\sfdefault}{bx}{n}

\usepackage{graphicx}
\usepackage{xcolor}
\usepackage{subcaption}
\usepackage{wrapfig}
\usepackage{needspace}
\usepackage{placeins}
\usepackage{listings}
\usepackage{tcolorbox}
\tcbuselibrary{breakable}
\usepackage{algorithm}
\usepackage{algpseudocode}
\usepackage{amssymb}
\usepackage{booktabs}
\usepackage{multirow}
\usepackage{url}
\usepackage{fancyhdr}
\usepackage{etoolbox}
\definecolor{darkblue}{rgb}{0,0.08,0.45}
\usepackage[colorlinks=true,linkcolor=darkblue,citecolor=darkblue,urlcolor=darkblue]{hyperref}

\newcounter{bibcount}
\makeatletter
\patchcmd{\@lbibitem}{\item[}{\item[\hfil\stepcounter{bibcount}{[\thebibcount]}}{}{}
\renewcommand\NAT@bibsetup[1]{%
  \setlength{\leftmargin}{\bibhang}%
  \setlength{\itemindent}{-\parindent}%
  \setlength{\itemsep}{\bibsep}%
  \setlength{\parsep}{\z@}}
\makeatother

\title{State-OPD: State-Grounded On-Policy Distillation \\ for Multi-Turn Agents}

\author{
\textbf{Zhentao Zhang}$^{1}$\thanks{Equal contribution.} \quad
\textbf{Yanzhao Zheng}$^{1}$\footnotemark[1] \quad
\textbf{Yanlin Liu}$^{1,2}$\footnotemark[1] \\[0.3em]
\textbf{Yuanqiang Yu}$^{1}$ \quad
\textbf{Tianze Xu}$^{1,3}$ \quad
\textbf{Chao Ma}$^{1}$ \quad
\textbf{Jihuai Zhu}$^{1}$ \quad
\textbf{Lelun Wu}$^{1,4}$ \\[0.3em]
\textbf{Baohua Dong}$^{1}$\thanks{Corresponding author.} \quad
\textbf{Hangcheng Zhu}$^{1}$ \quad
\textbf{Ruohui Huang}$^{1}$ \quad
\textbf{Gang Yu}$^{1}$ \\[0.8em]
\textsuperscript{1}Alibaba Group \quad
\textsuperscript{2}Tsinghua University \\
\textsuperscript{3}Shanghai Jiao Tong University \quad
\textsuperscript{4}University of Science and Technology of China \\[0.3em]
\normalfont\texttt{\{zhangzhentao.zzt,zhengyanzhao.zyz\}@alibaba-inc.com}
}

\newtheorem{proposition}{Proposition}

\begin{document}
\maketitle
\thispagestyle{plain}

\begin{abstract}
On-Policy Distillation (OPD) has emerged as a powerful paradigm for transferring reasoning capabilities from frontier models to smaller models, while extending its success to multi-turn agentic environments remains challenging. In this work, we investigate agentic OPD through \textit{interaction states}, distinguishing the objective task and executed action--observation record from the internal semantic trajectory of student reasoning. Through this lens, we identify two roles of student reasoning in multi-turn OPD: prior reasoning distorts teacher supervision, while current reasoning shapes the action alternatives sampled at the same state. Specifically, at matched interaction states, removing prior student reasoning improves teacher continuation success and changes the teacher's token-level targets, with larger continuation gains at later turns. Meanwhile, resampling complete reasoning–action responses yields greater action diversity at the same state than resampling actions under a fixed reasoning trace. Building on these findings, we introduce \textbf{State-Grounded On-Policy Distillation (State-OPD)}, a framework that grounds teacher supervision in interaction states and reallocates the sampling budget to generate multiple complete reasoning–action responses at states with high response entropy, while retaining full-history student generation. Comprehensive evaluations across ALFWorld, ScienceWorld and WebShop demonstrate that State-OPD achieves the highest macro-average success rate among the evaluated methods for both Qwen3-1.7B and Qwen3-4B students. Notably, for Qwen3-4B, State-OPD raises macro-average success from 26.07\% with the strongest baseline to 31.74\% and surpasses its Qwen3-32B teacher on ALFWorld Seen. Further analyses reveal that State-OPD achieves higher task success with fewer supervised turn responses and substantially reduces persistent failed-action repetition; qualitative traces illustrate decision revision based on interaction feedback.
\end{abstract}

\section{Introduction}
\label{sec:introduction}

On-policy distillation (OPD) has emerged as an effective paradigm for post-training language models by applying dense supervision from a stronger teacher to responses sampled from the student's own policy~\citep{agarwal2024policy,gu2024minillm,lu2025onpolicydistillation,jia2026aopd}.
Recent work has increasingly extended OPD to multi-turn agentic tasks, where language models make sequential decisions through interactions with external environments~\citep{wang2026tcod,li2026guidedopd,liao2026reopd,zhou2026turnopd,sun2026sopd,wang2026madopd}.
At each turn, the student generates a reasoning trace and an executable action, and the environment returns an observation~\citep{yao2022react,zhou2024lats}.
The accumulated history of reasoning, actions, and observations conditions both subsequent student decisions and teacher supervision.
Prior studies report that supervision can become unstable or less informative as interactions lengthen, making reliable knowledge transfer along student-generated trajectories challenging~\citep{wang2026tcod,zhou2026turnopd,fu2026revisitingopd}.

Existing methods address this challenge by reshaping rollout histories through curricula or teacher assistance~\citep{wang2026tcod,li2026guidedopd,liao2026reopd}, or by reallocating rollout and loss budgets across turns~\citep{zhou2026turnopd}.
Related analyses attribute this degradation to compounding rollout errors or shared-context compression~\citep{wang2026tcod,zhou2026turnopd}.
Complementary work on multi-turn dialogue distills history-conditioned student responses using references constructed from clean task evidence~\citep{lin2026ccopd,zheng2026maigo}.
In interactive environments, earlier assistant responses also contain executed actions that determine subsequent observations.
We therefore distinguish the realized action--observation record from the reasoning that accompanies it, and examine their roles in teacher supervision and student response sampling.

To separate the realized interaction from the student's interpretation of it, we define the interaction state $s_t$ as the task instruction and the sequence of executed actions and returned observations, excluding prior student reasoning.
Histories with different reasoning can therefore share the same $s_t$, allowing us to isolate reasoning's effect without changing the interaction that occurred.
Controlled interventions show that removing inherited student reasoning improves teacher continuation success from the same interaction states, with larger gains at later turns.
On identical student responses, removing inherited reasoning also reverses the decline of sampled reverse KL with turn depth, revealing its effect on the teacher's supervision.
On the student side, holding the pre-turn history fixed, resampling both current reasoning and action exposes a wider range of actions than resampling only the action after a fixed reasoning trace.
Together, these findings reveal an asymmetric role for student reasoning in agentic OPD: prior reasoning inherited by the teacher distorts supervision, while sampling current reasoning together with actions increases action diversity at the same interaction state.

Building on this asymmetry, we introduce \textbf{State-Grounded On-Policy Distillation (State-OPD)}, which organizes teacher supervision and student response sampling around interaction states.
State-grounded supervision conditions the teacher on the interaction state, while the student continues to generate responses from its full history.
State-centered response allocation reallocates the turn-response budget toward states with high response entropy and resamples complete reasoning–action responses to increase action diversity.
We evaluate State-OPD with two student-teacher pairs on three multi-turn agent benchmarks: ALFWorld~\citep{shridhar2020alfworld}, ScienceWorld~\citep{wang2022scienceworld}, and WebShop~\citep{yao2022webshop}.
Across five evaluation settings, State-OPD achieves the highest macro-average success rate for both student sizes, reaching 31.74\% with Qwen3-4B versus 26.07\% for the strongest baseline.
Notably, on ALFWorld Seen, the Qwen3-4B student surpasses its Qwen3-32B teacher, achieving 43.39\% success versus 38.39\%.
Training and behavioral analyses on ALFWorld further show sustained gains with fewer supervised turn responses and a substantial reduction in persistent failed-action loops.

Our main contributions are as follows:
\begin{itemize}
    \item We formalize interaction states in agentic OPD and reveal an asymmetric role of student reasoning: inherited reasoning distorts teacher supervision, while fixing a single current reasoning path limits the actions entering supervision.
    \item We introduce State-OPD, which makes interaction states the unit of agentic distillation: state-grounded teacher targets improve supervision quality, while entropy-guided complete-response resampling broadens the student actions entering supervision.
    \item We demonstrate leading macro-average success across three benchmarks and two student scales, with ALFWorld gains sustained using fewer supervised turn responses and far fewer persistent failed-action loops.
\end{itemize}

\section{Preliminaries}
\label{sec:preliminaries}

\paragraph{Multi-turn agent interaction.}
Given a task instruction $g$, an agent with policy $\pi_\theta$ interacts with an external environment over a finite horizon of $T$ turns.
At turn $t$, the agent observes $o_t$, conditions on the full interaction history
\begin{equation}
    h_t =
    (g,o_0,r_0,o_1,\ldots,r_{t-1},o_t),
    \label{eq:interaction-history}
\end{equation}
and generates a response
$r_t=(z_t,a_t)\sim\pi_\theta(\cdot\mid h_t)$, where $z_t$ denotes a reasoning trace and
$a_t=\psi(r_t)$ is the executable action extracted from the response.
The environment executes $a_t$ and returns the next observation $o_{t+1}$.
Repeating this interaction produces a trajectory
$\tau=(g,o_0,r_0,o_1,\ldots,r_{T-1},o_T)$,
which terminates after completion of the task or at the maximum horizon.

\paragraph{On-policy distillation for multi-turn agents.}
Let $\pi_\theta$ and $\pi_T$ denote the student policy and the teacher policy, respectively.
In multi-turn OPD, the student samples a trajectory $\tau\sim\pi_\theta$, and the teacher provides a target distribution over the student's response at each turn.
The response-level reverse-KL objective is
\begin{equation}
    \mathcal{L}_{\mathrm{OPD}}(\theta)
    =
    \mathbb{E}_{\tau\sim\pi_\theta}
    \left[
        \sum_{t=0}^{T-1}
        D_{\mathrm{KL}}
        \left(
            \pi_\theta(\cdot\mid h_t)
            \,\middle\|\,
            \pi_T(\cdot\mid h_t)
        \right)
    \right].
    \label{eq:agentic-opd}
\end{equation}

\begin{figure*}[t]
    \centering
    \includegraphics[width=\textwidth]{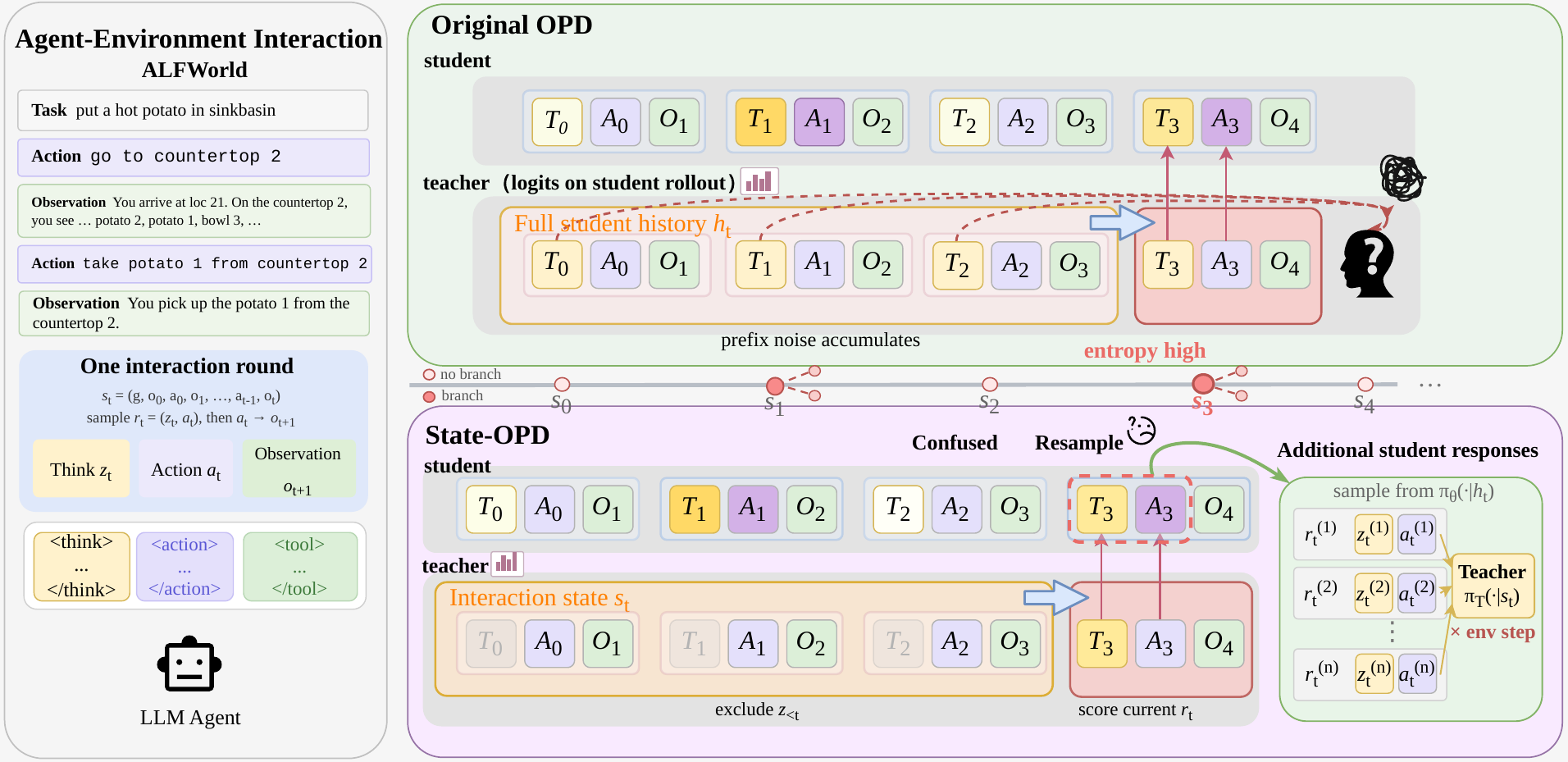}
    \caption[Overview of State-OPD.]{
    Overview of State-OPD. The teacher conditions on the interaction state rather than prior student reasoning; the student retains its full history and samples additional complete responses at states with high response entropy.
    }
    \label{fig:state_opd_pipeline}
\end{figure*}

\section{Method}
\label{sec:method}

\subsection{Interaction State as the Unit of Agentic Distillation}
\label{sec:state-separation}

Agentic OPD uses the student's accumulated history both to construct teacher targets and to sample subsequent student responses.
That history combines the realized action--observation record with the student's reasoning about it, making their effects on supervision difficult to distinguish.

We define the interaction state at turn $t$ as
\begin{equation}
    s_t =
    (g,o_0,a_0,o_1,\ldots,a_{t-1},o_t).
    \label{eq:interaction-state}
\end{equation}
The state $s_t$ retains the task instruction and the complete sequence of executed actions and returned observations while omitting $z_{<t}$.
Histories with different semantic trajectories therefore share the same $s_t$ when their executed actions and returned observations are identical.

\begin{proposition}[Interaction-state sufficiency]
\label{prop:interaction-sufficiency}
Let $\xi_t$ denote the latent environment state.
Suppose that each response affects the environment only through its executable
action $a_t=\psi(r_t)$ and that all environment feedback is recorded in the
observations.
Then, $s_t$ preserves all information in $h_t$ about $\xi_t$:
\begin{equation}
    p(\xi_t\mid h_t)=p(\xi_t\mid s_t).
    \label{eq:interaction-sufficiency}
\end{equation}
\end{proposition}

The proof is given in Appendix~\ref{app:interaction-sufficiency}.
Under the stated assumptions, $z_{<t}$ adds no information about the latent environment state $\xi_t$ beyond $s_t$, although it can still affect a teacher distribution conditioned on the full history.
This is an information-preservation result; whether removing prior reasoning improves a finite-capacity teacher is an empirical question.
In Eq.~\eqref{eq:agentic-opd}, prior and current reasoning enter through different routes: $z_{<t}$ conditions the teacher target, whereas $z_t$ shapes the actions in sampled student responses.
We examine how these dependencies affect teacher guidance and the diversity of student decisions brought under supervision at matched interaction states.

\subsection{Removing Inherited Reasoning Improves Teacher Guidance}
\label{sec:reasoning-history-interference}

\begin{figure*}[t]
    \centering
    \begin{subfigure}[t]{0.335\linewidth}
        \centering
        \includegraphics[width=\linewidth]{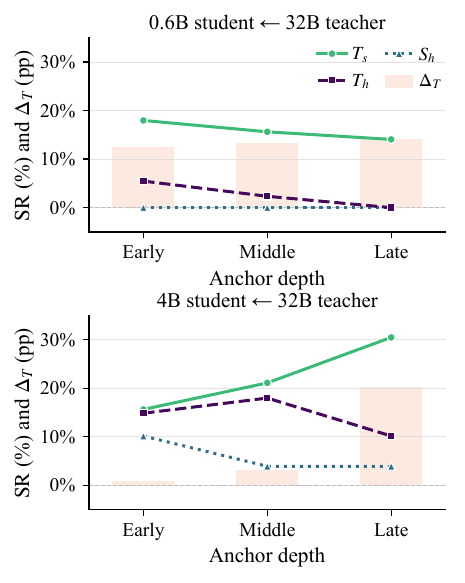}
        \caption{Teacher continuation success. At matched interaction states, $T_h$ uses the student history, $T_s$ uses the interaction state, and $S_h$ is the student reference. Shaded bars show $\Delta_T(d)$.}
        \label{fig:history_recovery}
    \end{subfigure}
    \hfill
    \begin{subfigure}[t]{0.64\linewidth}
        \centering
        \includegraphics[width=\linewidth]{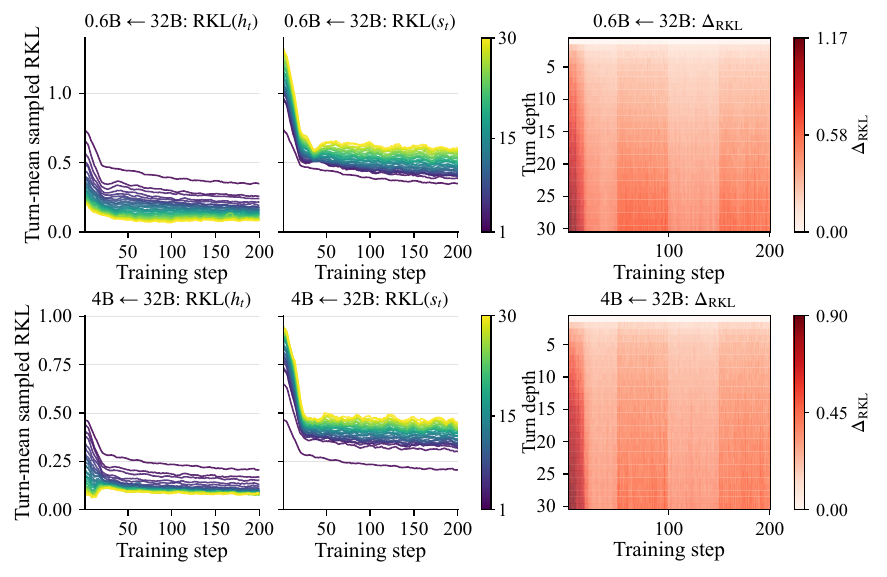}
        \caption{Sampled reverse KL by teacher context and turn depth. Curves progress from early (purple) to late (yellow) turns; heatmaps show $\Delta_{\mathrm{RKL}}=\mathrm{RKL}(s_t)-\mathrm{RKL}(h_t)$.}
        \label{fig:history_rkl}
    \end{subfigure}
    \caption{Removing prior student reasoning improves Qwen3-32B teacher continuation at ALFWorld states visited by Qwen3-0.6B and Qwen3-4B students and changes the reverse-KL depth profile. Continuations run for at most 30 additional turns. Paired intervals appear in Table~\ref{tab:teacher-continuation-details}.}
    \label{fig:history_intervention}
\end{figure*}

At a fixed interaction state, the teacher already has the complete record of executed actions and environmental feedback.
By Proposition~\ref{prop:interaction-sufficiency}, prior student reasoning adds no environment-state information beyond $s_t$, yet vanilla OPD includes it in the teacher context used to form the supervision target.
We assess teacher guidance through paired continuations and characterize teacher targets through fixed-response comparisons at the same $s_t$.

\paragraph{Analysis setup.}
We select non-terminal anchors from base-student trajectories on ALFWorld, using Qwen3-0.6B and Qwen3-4B students and a Qwen3-32B teacher.
At each anchor, we replay the executed action prefix to restore the environment and compare teacher continuations conditioned on $h_t$ or $s_t$, denoted $T_h$ and $T_s$.
We also report the full-history student continuation $S_h$ as a reference.
Each $T_h/T_s$ pair shares $s_t$, the restored environment, and the sampling seed; the intervention removes only $z_{<t}$ from the teacher's inherited context.
Subsequent responses and observations are appended in full in both conditions.

We group anchors into three equal depth ranges and measure the paired difference in teacher continuation success at depth $d$ as
\begin{equation}
    \Delta_T(d)
    =
    \operatorname{SR}(T_s,d)
    -
    \operatorname{SR}(T_h,d),
    \label{eq:teacher-recovery-gap}
\end{equation}
where a positive value measures the gain from removing preceding student reasoning at the same interaction states.
Appendix~\ref{app:history-interference-details} provides the complete protocols for the continuation and reverse-KL analyses.

\paragraph{Teacher guidance improves without inherited reasoning.}
As illustrated in Figure~\ref{fig:history_intervention}\subref{fig:history_recovery}, $T_s$ achieves higher continuation success than $T_h$ for both student models in every anchor range.
The effect is especially pronounced for Qwen3-4B, where the paired gain $\Delta_T(d)$ rises from 0.8 percentage points at early anchors to 20.3 points at late anchors.
The growing $\Delta_T(d)$ indicates that inherited student reasoning increasingly weakens the teacher's ability to guide the task as interactions deepen.

\paragraph{Inherited reasoning distorts the distillation target.}
To test whether this context effect also enters OPD supervision, we run vanilla OPD for 200 optimizer steps and evaluate each sampled response under two teacher contexts.
The student context, response tokens, and student log probabilities remain fixed; only the teacher's preceding context changes between $h_t$ and $s_t$.
Figure~\ref{fig:history_intervention}\subref{fig:history_rkl} shows that sampled reverse KL decreases with turn depth under $h_t$-conditioning, as reported in prior agentic OPD analyses~\citep{zhou2026turnopd}.
Under $s_t$-conditioning, the trend reverses; the paired heatmaps show this context-dependent difference across training.
Together, these results suggest that inherited reasoning pulls the full-history teacher target toward the student's response distribution at later turns: sampled KL falls even as the teacher provides weaker task-level guidance.

\subsection{A Single Reasoning Path Limits Action Diversity}
\label{sec:state-local-coverage}

\begin{figure*}[t]
    \centering
    \begin{subfigure}[t]{0.485\linewidth}
        \centering
        \includegraphics[width=\linewidth]{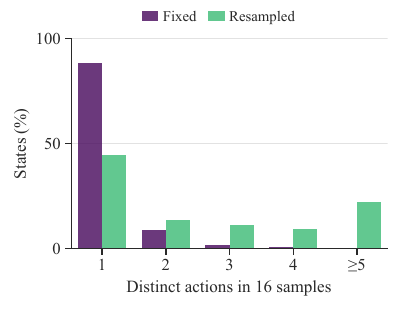}
        \caption{Action diversity by reasoning mode.}
        \label{fig:state-response-space}
    \end{subfigure}
    \hfill
    \begin{subfigure}[t]{0.485\linewidth}
        \centering
        \includegraphics[width=\linewidth]{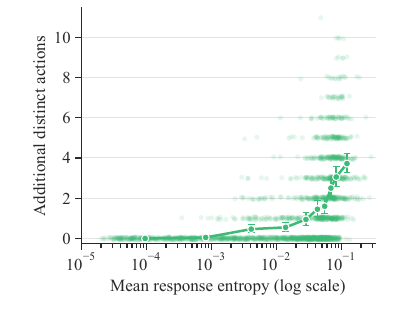}
        \caption{Response entropy and additional actions.}
        \label{fig:state-entropy-coverage}
    \end{subfigure}
    \caption{Resampling reasoning reveals more actions at the same state. (a) For each state, we draw 16 responses with fixed reasoning and 16 complete responses. Side-by-side bars show the percentage of states in each action-count bin within each condition. (b) Faint points show individual states; solid markers show equal-count-bin means with 95\% trajectory-bootstrap intervals. Higher response entropy accompanies more additional actions.}
    \label{fig:state-coverage}
\end{figure*}

In each student response $r_t=(z_t,a_t)$, the sampled reasoning path precedes and shapes the sampled action.
The loss supervises every response token, but the realized actions represented in an update still depend on which reasoning paths were sampled.
We test whether sampling complete responses from the same pre-turn history $h_t$ reveals actions missed when $z_t$ is fixed.

For the reasoning-path probe, we collect 64 Qwen3-4B trajectories on ALFWorld, each with up to 30 interaction turns.
At each visited turn, we hold $h_t$ fixed and construct two matched 16-response sets, each containing the realized response and 15 additional samples.
Complete-response resampling regenerates both reasoning and action, whereas fixed-reasoning resampling retains the realized reasoning through \texttt{</think>} and regenerates only the action suffix.
For the entropy--diversity probe, we use a separate batch of 64 trajectories and sample 16 complete responses at each visited state.
We measure action diversity by counting distinct valid parsed actions after whitespace trimming and lowercasing, as detailed in Appendix~\ref{app:state-coverage-details}.

Figure~\ref{fig:state-coverage}\subref{fig:state-response-space} shows that 88.5\% of fixed-reasoning probes yield just one parsed action, compared with 44.3\% of complete-response probes; the latter more often yield multiple distinct actions.
Varying current reasoning thus exposes action alternatives that sampling only the action suffix misses at the same interaction state.
Figure~\ref{fig:state-coverage}\subref{fig:state-entropy-coverage} further shows that higher response entropy is associated with more additional distinct actions.
Together, these probes motivate sampling complete responses and allocating additional samples to states where they are more likely to reveal new student decisions.

\subsection{State-OPD: Distillation over Interaction States}
\label{sec:state-opd}

Building on these findings, we introduce \textsc{State-OPD}, which distills multiple complete student responses at selected interaction states against teacher targets grounded in the corresponding action--observation records, as Figure~\ref{fig:state_opd_pipeline} illustrates.
This lets teacher guidance follow the interaction the agent actually reached and shape a wider range of its possible next decisions, sustaining reliable knowledge transfer as student-generated trajectories grow longer.

\paragraph{State-grounded supervision.}
Let $\mathcal{R}$ denote the indexed response pool for an update, with each entry $(i,t,r)$ associating a student response $r$ with task $i$ and visited turn $t$.
For each $(i,t,r)\in\mathcal{R}$, State-OPD evaluates the student under the full history $h_{i,t}$ and the teacher under the interaction state $s_{i,t}$, while conditioning both distributions on the same realized response prefix $r_{<k}$:
\begin{equation}
    \mathcal{L}_{\mathrm{State\text{-}OPD}}(\theta)
    =
    \frac{
        \displaystyle
        \sum_{(i,t,r)\in\mathcal{R}}
        \sum_{k=1}^{|r|}
        D_{\mathrm{KL}}\!\left(
            \pi_\theta(\cdot\mid h_{i,t},r_{<k})
            \,\middle\|\,
            \pi_T(\cdot\mid s_{i,t},r_{<k})
        \right)
    }{
        \displaystyle
        \sum_{(i,t,r)\in\mathcal{R}} |r|
    }.
    \label{eq:state-opd-objective}
\end{equation}
State grounding therefore removes the preceding student reasoning $z_{<t}$ only from the teacher's context; the student's on-policy generation context and the current autoregressive response prefix remain unchanged.
As detailed in Appendix~\ref{app:sampled-opd-update}, the experiments use per-token teacher--student log-probability advantages in a clipped policy surrogate.

\paragraph{State-centered response allocation.}
To bring more reasoning-induced actions under supervision within a fixed nominal turn-response budget, we concentrate complete-response samples at selected interaction states.
We retain every root-trajectory response in $\mathcal{R}$ and expand the pool at selected states by regenerating both reasoning and action from the same $h_{i,t}$.
For each root response $r_{i,t}$ of length $L_{i,t}$, we compute the student's full-vocabulary token entropy averaged over its realized tokens,
\begin{equation}
    \overline{\mathcal{H}}_{i,t}
    =
    \frac{1}{L_{i,t}}
    \sum_{k=1}^{L_{i,t}}
    \mathcal{H}\!\left[
        \pi_\theta\!\left(\cdot \mid h_{i,t},r_{i,t,<k}\right)
    \right].
    \label{eq:response-entropy}
\end{equation}
Let $M$ denote the rollout multiplicity of the reference OPD configuration and $\beta\in\mathbb{N}_{+}$ a budget-concentration factor.
For a root trajectory with $T_i$ turns, we select the $P_i$ highest-entropy states and sample $n$ additional complete responses at each:
\begin{equation}
\begin{aligned}
    n &= \beta(M-1), \\
    P_i &= \left\lceil\frac{T_i}{\beta}\right\rceil.
\end{aligned}
    \label{eq:state-expansion-budget}
\end{equation}
Up to integer rounding, $T_i+nP_i\approx MT_i$, matching the nominal turn-response budget of the $M$-rollout reference; $\beta$ reallocates this budget toward more responses at fewer states.
The additional responses enter Eq.~\eqref{eq:state-opd-objective}, but their actions are not executed and produce no alternative environment suffix.
The complete training procedure is given in Algorithm~\ref{alg:state-opd}.

\section{Experiments}
\label{sec:experiments}

\subsection{Experimental Setup}
\label{sec:experimental-setup}

\begin{figure*}[!t]
  \begin{minipage}{\textwidth}
    \centering
    \captionof{table}{Main results on ALFWorld, ScienceWorld, and WebShop. SR is success rate (\%); Rounds is the mean interaction count, assigning failures the maximum horizon; Avg. is the macro-average over five evaluation settings. Bold and underlining mark the best and second-best trained methods within each student group.}
    \label{tab:main-results}
    \footnotesize
    \setlength{\tabcolsep}{2.5pt}
    \renewcommand{\arraystretch}{1.08}
    \resizebox{\textwidth}{!}{%
    \begin{tabular}{l*{10}{c}cc}
        \toprule
        \multirow{3}{*}{Method}
        & \multicolumn{4}{c}{ALFWorld}
        & \multicolumn{4}{c}{ScienceWorld}
        & \multicolumn{2}{c}{\multirow{2}{*}{WebShop}}
        & \multicolumn{2}{c}{\multirow{2}{*}{Avg.}} \\
        \cmidrule(lr){2-5}\cmidrule(lr){6-9}
        &
        \multicolumn{2}{c}{Seen}
        & \multicolumn{2}{c}{Unseen}
        & \multicolumn{2}{c}{Seen}
        & \multicolumn{2}{c}{Unseen}
        & & & & \\
        \cmidrule(lr){2-3}\cmidrule(lr){4-5}\cmidrule(lr){6-7}\cmidrule(lr){8-9}\cmidrule(lr){10-11}\cmidrule(lr){12-13}
        &
        SR $\uparrow$ & Rounds $\downarrow$
        & SR $\uparrow$ & Rounds $\downarrow$
        & SR $\uparrow$ & Rounds $\downarrow$
        & SR $\uparrow$ & Rounds $\downarrow$
        & SR $\uparrow$ & Rounds $\downarrow$
        & SR $\uparrow$ & Rounds $\downarrow$ \\
        \midrule
        \textbf{Qwen3-32B (Teacher)}
        & 38.39 & 22.98 & 46.27 & 22.76 & 31.58 & 25.17 & 30.95 & 23.87 & 19.25 & 12.89 & 33.29 & 21.54 \\
        \midrule
        \multicolumn{13}{c}{\textbf{Qwen3-1.7B (Student)}} \\
        \addlinespace[1pt]
        Zero-shot
        & 2.32 & 29.64 & 3.36 & 29.43 & 1.32 & 29.66 & 0.90 & 29.78 & 10.35 & 13.81 & 3.65 & 26.47 \\
        Vanilla OPD
        & 26.43 & 25.42 & 32.46 & 24.74 & 10.53 & 27.95 & \underline{13.17} & \textbf{27.00} & 11.90 & 13.60 & 18.90 & 23.74 \\
        TCOD-F2B
        & \underline{29.11} & \underline{24.83} & 28.54 & 25.12 & 9.65 & 28.15 & 11.13 & 27.54 & 11.40 & 13.66 & 17.97 & 23.86 \\
        Guided-OPD
        & 27.86 & 25.11 & 35.45 & \underline{24.11} & \textbf{13.38} & \textbf{27.39} & \textbf{13.30} & \underline{27.08} & 12.15 & \underline{13.59} & \underline{20.43} & \underline{23.46} \\
        TurnOPD
        & 27.32 & 25.22 & \textbf{37.69} & 24.31 & 8.33 & 28.21 & 10.49 & 27.55 & \textbf{12.80} & \textbf{13.51} & 19.33 & 23.76 \\
        \cmidrule(lr){1-13}
        State-OPD
        & \textbf{31.25} & \textbf{24.52} & \underline{36.01} & \textbf{23.90} & \underline{11.18} & \underline{27.71} & \underline{13.17} & 27.13 & \underline{12.20} & 13.60 & \textbf{20.76} & \textbf{23.37} \\
        \midrule
        \multicolumn{13}{c}{\textbf{Qwen3-4B (Student)}} \\
        \addlinespace[1pt]
        Zero-shot
        & 13.75 & 27.68 & 14.55 & 27.53 & 11.18 & 27.92 & 14.45 & 26.98 & 15.95 & 13.21 & 13.98 & 24.66 \\
        Vanilla OPD
        & \underline{30.89} & \underline{24.38} & 33.96 & 24.53 & 23.25 & 26.00 & \underline{25.70} & 24.54 & 16.55 & 13.12 & \underline{26.07} & 22.51 \\
        TCOD-F2B
        & 29.29 & 24.77 & \underline{35.26} & \underline{24.05} & 23.46 & 25.78 & 24.94 & 24.54 & 15.40 & 13.24 & 25.67 & \underline{22.47} \\
        Guided-OPD
        & 27.14 & 24.91 & 30.78 & 24.83 & \textbf{25.22} & \textbf{25.59} & \textbf{26.34} & \textbf{24.16} & 16.55 & 13.11 & 25.21 & 22.52 \\
        TurnOPD
        & 26.79 & 25.05 & 33.58 & 24.46 & 21.71 & 25.96 & 25.45 & \underline{24.45} & \underline{17.80} & \underline{13.00} & 25.07 & 22.58 \\
        \cmidrule(lr){1-13}
        State-OPD
        & \textbf{43.39} & \textbf{22.95} & \textbf{45.71} & \textbf{23.09} & \underline{24.78} & \underline{25.73} & 25.19 & 24.73 & \textbf{19.65} & \textbf{12.82} & \textbf{31.74} & \textbf{21.86} \\
        \bottomrule
    \end{tabular}
    }
  \end{minipage}
  \par\vspace{0.5\baselineskip}
    \nextfloat
    \centering
    \includegraphics[width=0.27\textwidth]{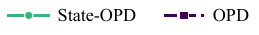}\\[-0.4em]
    \begin{subfigure}[t]{0.238\textwidth}
        \centering
        \includegraphics[width=\linewidth]{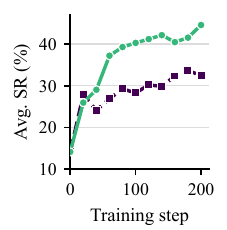}
        \caption{Success rate}
        \label{fig:alfworld-training-sr}
    \end{subfigure}\hfill
    \begin{subfigure}[t]{0.238\textwidth}
        \centering
        \includegraphics[width=\linewidth]{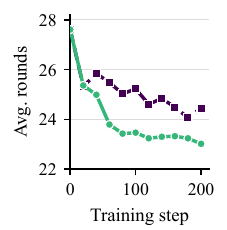}
        \caption{Interaction rounds}
        \label{fig:alfworld-training-rounds}
    \end{subfigure}\hfill
    \begin{subfigure}[t]{0.238\textwidth}
        \centering
        \includegraphics[width=\linewidth]{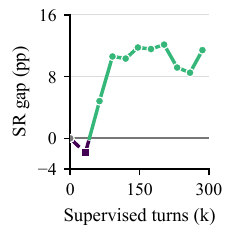}
        \caption{Supervision efficiency}
        \label{fig:alfworld-training-supervision}
    \end{subfigure}\hfill
    \begin{subfigure}[t]{0.238\textwidth}
        \centering
        \includegraphics[width=\linewidth]{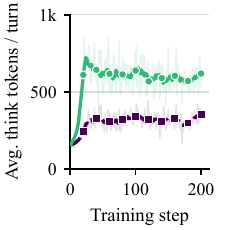}
        \caption{Think tokens / turn}
        \label{fig:alfworld-training-length}
    \end{subfigure}
    \caption{Training dynamics on ALFWorld for a Qwen3-4B student distilled from Qwen3-32B. (a,b) Seen/Unseen averages. (c) Success-rate gap at matched supervised-turn budgets (State-OPD minus linearly interpolated OPD). (d) Mean thinking tokens per root-trajectory turn; faint lines show step values and marked lines show centered 11-step moving averages.}
    \label{fig:alfworld-training-dynamics}
\end{figure*}

\paragraph{Implementation.}
We distill Qwen3-1.7B and Qwen3-4B students from a frozen Qwen3-32B teacher~\citep{yang2025qwen3}.
All methods use a batch size of 16 tasks. Each trained baseline samples four rollouts per task. State-OPD uses the corresponding reference multiplicity $M=4$ and fixes $\beta=2$ across environments and model sizes, allocating six additional responses to each state in the highest-entropy half of the root trajectory while approximately matching the baseline turn-response supervision budget.
We implement all methods in veRL and run each experiment on 32 GPUs with 96\,GB of memory each. Each configuration uses one training seed (42). Further training and evaluation configurations are provided in Appendix~\ref{app:training-details}.

\paragraph{Benchmarks.}
We evaluate experiments on three benchmarks: ALFWorld~\citep{shridhar2020alfworld} for embodied household tasks, ScienceWorld~\citep{wang2022scienceworld} for interactive scientific reasoning, and WebShop~\citep{yao2022webshop} for web-based shopping.
We use four evaluation rollouts per task for ALFWorld and WebShop and one for ScienceWorld.
We report success rate (SR $\uparrow$) and average interaction rounds (Rounds $\downarrow$), with unsuccessful episodes counted at the corresponding environment's maximum horizon.
More benchmark and evaluation details are provided in Appendix~\ref{app:benchmark-details}.

\paragraph{Baselines.}
We compare State-OPD with four primary methods: Vanilla OPD, TCOD-F2B~\citep{wang2026tcod}, Guided-OPD~\citep{li2026guidedopd}, and TurnOPD~\citep{zhou2026turnopd}.
We additionally report the zero-shot students and the frozen teacher as reference policies.
Full implementation details are in Appendix~\ref{app:baseline-implementations}.

\subsection{Main Results}
\label{sec:main-results}

\Needspace{10\baselineskip}
\paragraph{State-OPD achieves the highest observed macro-average success.}
Table~\ref{tab:main-results} shows that State-OPD achieves the highest macro-average success rate and fewest average interaction rounds among trained methods for both Qwen3-1.7B and Qwen3-4B students.
For Qwen3-4B, State-OPD raises average success from 26.07\% with the strongest baseline, Vanilla OPD, to 31.74\%; the Qwen3-1.7B student also leads its group at 20.76\%.
The largest gains occur on ALFWorld, where the 4B student reaches 43.39\% success on Seen and 45.71\% on Unseen, surpassing the strongest baselines, Vanilla OPD on Seen and TCOD-F2B on Unseen, by 12.50 and 10.45 percentage points, respectively.
Table~\ref{tab:paired-bootstrap-sr} reports the corresponding task-paired 95\% confidence intervals of $[8.39,16.79]$ and $[5.04,15.86]$ percentage points, conditional on the fixed trained checkpoints.
Notably, this 4B student outperforms the Qwen3-32B teacher on ALFWorld Seen, achieving 43.39\% versus 38.39\% success despite having one-eighth as many parameters.

\paragraph{Sustained gains with fewer supervised turn responses.}
Figures~\ref{fig:alfworld-training-sr} and~\ref{fig:alfworld-training-rounds} show that State-OPD's ALFWorld advantage persists through training: from step 40 onward, the Qwen3-4B student has higher success and fewer interaction rounds than Vanilla OPD at every evaluation checkpoint.
At step 200, it reaches 44.55\% success in 23.02 rounds, compared with 32.42\% success in 24.46 rounds for Vanilla OPD.
Figure~\ref{fig:alfworld-training-supervision} shows the success-rate gap at matched supervision budgets: after 146k supervised turn responses, State-OPD reaches 40.26\% success, 11.77 percentage points above Vanilla OPD interpolated to the same budget.
State-OPD thus sustains its ALFWorld advantage across training and finishes after supervising approximately 286k turn responses, compared with 302k for Vanilla OPD.

\paragraph{State-OPD revisits interaction evidence and breaks failed-action loops.}
Figure~\ref{fig:alfworld-training-length} shows that State-OPD produces substantially longer thinking responses than Vanilla OPD, reaching 638 versus 336 thinking tokens per root-trajectory turn at step 200.
In the task-matched ALFWorld trace in Appendix~\ref{app:state-review-case}, the State-OPD student reconstructs the locations it has searched and the observations obtained there before choosing its next search action.
Although the student's full history $h_t$ remains available at evaluation, its reasoning in this trace reassesses the executed interaction before acting.

\noindent\begin{minipage}[t]{0.41\textwidth}
    \vspace{0pt}
    \centering
    \captionsetup{type=figure}
    \includegraphics[width=\linewidth]{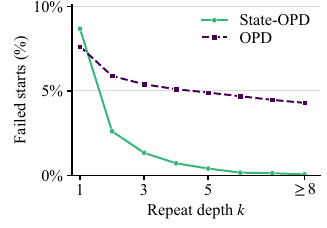}
    \caption{State-OPD reduces persistent failed-action loops on ALFWorld. At step 200 for a Qwen3-4B student distilled from Qwen3-32B, the curves show the share of failed-action starts with at least $k$ further identical actions, each returning \texttt{Nothing happens.}}
    \label{fig:failed-action-repeat-depth}
\end{minipage}\hfill
\begin{minipage}[t]{0.56\textwidth}
\vspace{0pt}
State-OPD is substantially less likely to persist with an action that the environment has already rejected.
At step 200, Figure~\ref{fig:failed-action-repeat-depth} shows similar rates of one ineffective retry (8.67\% versus 7.63\% of failed-action starts), yet only 0.07\% of State-OPD failure starts reach eight further identical failures, compared with 4.29\% for Vanilla OPD.
Figure~\ref{fig:failed-action-repeat-training} extends this comparison across training: from step 40 onward, fewer than 0.9\% of State-OPD failure starts reach five further repeats, versus more than 4.9\% for Vanilla OPD at every checkpoint.
The task-matched failed-pickup and navigation traces in Appendix~\ref{app:reasoning-case-study} illustrate the corresponding decision-level behavior: State-OPD revises its next action after negative feedback, whereas Vanilla OPD repeats the rejected action.
The failure-loop advantage thus emerges after feedback arrives: State-OPD more readily turns an observation that contradicts its previous decision into a different action.
\end{minipage}\par

\FloatBarrier
\subsection{Ablation Studies}
\label{sec:ablation-studies}

We conduct a two-factor ablation on ALFWorld with a Qwen3-4B student and Qwen3-32B teacher, testing each component with and without the other.
Without state-centered response allocation, we retain Vanilla OPD's four full rollouts per task and change only the teacher context.

\begin{table}[!htbp]
    \centering
    \caption{State-OPD ablation on ALFWorld. A Qwen3-4B student is distilled from Qwen3-32B; Avg. is the mean of Seen and Unseen SR.}
    \label{tab:component-ablation}
    \footnotesize
    \setlength{\tabcolsep}{10pt}
    \renewcommand{\arraystretch}{1.10}
    \begin{tabular}{lccc}
        \toprule
        Method & Seen SR $\uparrow$ & Unseen SR $\uparrow$ & Avg. SR $\uparrow$ \\
        \midrule
        \textbf{State-OPD (Ours)}
        & \textbf{43.39} & \textbf{45.71} & \textbf{44.55} \\
        \midrule
        w/o State-Centered Response Allocation
        & 40.36 & 43.10 & 41.73 \\
        w/o State-Grounded Supervision
        & 33.39 & 37.31 & 35.35 \\
        Vanilla OPD
        & 30.89 & 33.96 & 32.42 \\
        \bottomrule
    \end{tabular}
\end{table}

Table~\ref{tab:component-ablation} shows that each component improves success in both settings of the other.
State grounding raises average SR from 32.42\% to 41.73\% without response allocation and from 35.35\% to 44.55\% with it.
Response allocation raises average SR from 32.42\% to 35.35\% under full-history teacher conditioning and from 41.73\% to 44.55\% under state-grounded conditioning.
With both components, State-OPD reaches 43.39\% Seen and 45.71\% Unseen SR, for a 44.55\% average, 12.13 points above Vanilla OPD.
Together with the preceding analyses, these ablations give a coherent account of State-OPD's gains: state grounding conditions the teacher target on $s_t$, response allocation brings more reasoning-induced actions at the same state into supervision, and their combination achieves the highest success on ALFWorld.

\paragraph{Sensitivity to response concentration.}
Across $\beta\in\{2,3,4\}$, State-OPD outperforms Vanilla OPD on both ALFWorld splits.
Figure~\ref{fig:allocation-analysis} in Appendix~\ref{app:allocation-analysis} also shows that entropy-ranked state selection recovers more distinct actions than random selection at every tested $\beta$ under matched budgets.

\section{Related Work}
\label{sec:related-work}

\paragraph{On-policy distillation.}
On-policy distillation trains on student-generated responses with dense teacher feedback~\citep{gu2024minillm,agarwal2024policy,lu2025onpolicydistillation}.
Subsequent work changes the token-level distillation objective or concentrates supervision on reasoning prefixes~\citep{jin2026entropyawareonpolicydistillationlanguage,jia2026aopd,zhang2026fasteffectiveonpolicydistillation,ko2026scalingreasoningefficientlyrelaxed,fu2026revisitingopd}.
Studies of data dependence also report substantial gains from a single training example or teacher-generated prompts~\citep{fu2026oneshotopd,li2026datafreeopd}.
Other analyses examine teacher--student reasoning compatibility and how the teacher's conditioning context, including privileged information, historical solutions, or peer rollouts, shapes supervision~\citep{li2026rethinkingonpolicydistillationlarge,kaur2026rethinkingonpolicyselfdistillationthinking,ye2026opcd,yu2026multirolloutopd}.
Context-cleaned dialogue methods address contamination from prior assistant replies: MAIGO removes those replies from middle-turn reference contexts~\citep{zheng2026maigo}, while CCOPD supervises final answers with a teacher conditioned on complete user evidence~\citep{lin2026ccopd}.
In agentic interaction, prior replies also contain executed actions that determine subsequent observations; State-OPD preserves this action--observation record while separating out inherited reasoning.

\paragraph{Agentic on-policy distillation.}
In interactive agents, student actions change subsequent observations, so both the rollout trajectory and the placement of supervision matter.
Existing methods shape the trajectories used for distillation through temporal curricula, teacher-guided turns, prefix replay, or adaptive stopping and restarting~\citep{wang2026tcod,li2026guidedopd,liao2026reopd,gui2026stride}.
Complementary methods vary supervision across turns or steps using turn-aware budgets, step-level corrections, divergence- or confidence-based weighting, and validation of teacher interventions through subsequent student rollouts~\citep{zhou2026turnopd,zhong2026sod,zhou2026sageopd,chen2026futurebridge,sun2026sopd,tan2026atod}.
State-alignment methods address mismatches between supervision and the student's rollout: SMRC-SD routes privileged guidance to states supported by successful reference trajectories~\citep{liu2026smrcsd}, while MemOPD reconstructs the original model-invocation context for scoring after memory updates~\citep{liu2026memopd}.
State-OPD identifies reasoning's asymmetric roles at a fixed interaction state: inherited reasoning distorts teacher targets, while resampling current reasoning together with actions increases action diversity.
This asymmetry motivates using interaction states to organize both teacher conditioning and entropy-guided complete-response sampling.

\section{Conclusion}
\label{sec:conclusion}

Analyzing agentic OPD at fixed interaction states reveals an asymmetry: inherited student reasoning weakens teacher guidance, while resampling current reasoning increases action diversity.
\textsc{State-OPD} grounds teacher targets in executed actions and observations and resamples complete responses at high-entropy states, while retaining full-history student generation.
Across three benchmarks, State-OPD achieves the highest macro-average success among trained methods at both student scales; its Qwen3-4B student also surpasses the Qwen3-32B teacher on ALFWorld Seen.
On ALFWorld, it sustains higher success with fewer supervised turn responses and sharply reduces persistent failed-action loops.
These findings establish interaction states as the organizing unit for agentic knowledge transfer.
Separating the realized interaction from the student's reasoning about it provides a common foundation for grounding teacher supervision and exploring alternative decisions at the same state.

\bibliography{main}
\bibliographystyle{related}

\clearpage
\appendix
\section{Proof of Interaction Sufficiency}
\label{app:interaction-sufficiency}

\paragraph{Proof of Proposition~\ref{prop:interaction-sufficiency}.}
Let $\xi_i$ denote the latent environment state at turn $i$.
The assumptions of Proposition~\ref{prop:interaction-sufficiency} yield the following factorization for a reachable full history $h_t$:
\begin{equation}
\begin{split}
    p(\xi_{0:t},h_t)
    ={}&p(g,\xi_0,o_0)
    \prod_{i=0}^{t-1}
    \pi_\theta(r_i\mid h_i)\\
    &{}\times
    P(\xi_{i+1},o_{i+1}\mid \xi_i,\psi(r_i)).
\end{split}
\label{eq:interaction-factorization}
\end{equation}
For a fixed $h_t$, write the right-hand side as
$C_\theta(h_t)F(\xi_{0:t},s_t)$, where
$C_\theta(h_t)=\prod_{i<t}\pi_\theta(r_i\mid h_i)$ contains the probability of the student-generated responses.
Because the environment receives each response only through
$a_i=\psi(r_i)$, the factor $F$ depends on the transcript only through
$s_t$.
$C_\theta(h_t)$ is constant with respect to the latent state path and therefore cancels when Equation~\eqref{eq:interaction-factorization} is normalized over $\xi_{0:t}$:
\begin{equation}
    p(\xi_{0:t}\mid h_t)
    \propto F(\xi_{0:t},s_t).
    \label{eq:full-history-posterior}
\end{equation}

Conditioning on $s_t$ instead sums over all full histories in its equivalence class:
\begin{equation}
\begin{split}
    p(\xi_{0:t},s_t)
    &{}=
    \sum_{\substack{h_t\text{ consistent}\\\text{with }s_t}}
    p(\xi_{0:t},h_t)\\
    &{}=
    F(\xi_{0:t},s_t)
    \sum_{\substack{h_t\text{ consistent}\\\text{with }s_t}}C_\theta(h_t).
\end{split}
\label{eq:state-posterior-factorization}
\end{equation}
The final sum does not depend on $\xi_{0:t}$ and likewise cancels under normalization.
Equations~\eqref{eq:full-history-posterior} and~\eqref{eq:state-posterior-factorization} therefore define the same posterior over the latent state path.
Marginalizing $\xi_{0:t-1}$ gives
$p(\xi_t\mid h_t)=p(\xi_t\mid s_t)$, which proves
Proposition~\ref{prop:interaction-sufficiency}.

\Needspace{18\baselineskip}
\section{State-OPD Training Procedure}
\label{app:training-procedure}

\begin{algorithm}[H]
\caption{State-Grounded On-Policy Distillation (\textsc{State-OPD})}
\label{alg:state-opd}
\begin{algorithmic}[1]
\Require $\pi_\theta$, $\pi_T$, $\mathcal{E}$, reference rollout multiplicity $M$, and concentration $\beta$
\For{each training iteration}
    \State Initialize the response pool $\mathcal{R}\gets\varnothing$
    \For{each sampled task instance $i$}
        \State Roll out $\tau_i\sim\pi_\theta$ in $\mathcal{E}$; add its turn responses to $\mathcal{R}$
        \State Compute $\overline{\mathcal{H}}_{i,t}$ for all turns using Eq.~\ref{eq:response-entropy}
        \State Select the top-$P_i$ states using Eq.~\ref{eq:state-expansion-budget}
        \State Add $n$ samples from $\pi_\theta(\cdot\mid h_{i,t})$ for every selected $t$
    \EndFor
    \State Score every $(i,t,r)\in\mathcal{R}$ with $\pi_T$ conditioned on $s_{i,t}$
    \State Update $\pi_\theta$ with the sampled policy surrogate in Appendix~\ref{app:sampled-opd-update}
\EndFor
\end{algorithmic}
\end{algorithm}

\section{Diagnostic Analysis Details}
\label{app:diagnostic-details}

\subsection{Reasoning-History Interference}
\label{app:history-interference-details}

\paragraph{Paired continuation intervention.}
We sample 64 ALFWorld source trajectories from each base student with temperature $1.0$, top-$p$ $1.0$, and top-$k$ $-1$, and divide the 30-turn horizon into early (turns 1--10), middle (turns 11--20), and late (turns 21--30) ranges.
Within each range, we retain at most 32 non-terminal anchors and at most three anchors from any source episode.
At each anchor, we restore the environment by replaying the executed action prefix and generate teacher continuations under $h_t$ and $s_t$, together with a full-history student continuation.
The teacher and student branches use the same base sampling seed for each anchor and repeat.
We draw four continuations per branch with temperature $1.0$, top-$p$ $1.0$, and top-$k$ $-1$.
Each continuation runs for at most 30 additional turns; each generated turn is capped at 512 tokens.
We compute context effects within matched anchor--repeat pairs.
Confidence intervals are obtained from 10,000 bootstrap resamples with seed 42, clustered by source episode so that all anchors and repeats from an episode remain in the same resampled cluster.

\begin{table}[!htbp]
    \centering
    \caption{Teacher continuation from matched ALFWorld anchors. Each row uses 32 anchors from 30 source episodes and four repeats per condition (128 paired continuations). $T_h$ and $T_s$ are success rates (\%); $\Delta_T$ is their paired difference in percentage points with a 95\% source-episode bootstrap interval.}
    \label{tab:teacher-continuation-details}
    \small
    \setlength{\tabcolsep}{5pt}
    \begin{tabular}{llrrl}
        \toprule
        Student & Anchor depth & $T_h$ & $T_s$ & $\Delta_T$ [95\% CI] \\
        \midrule
        \multirow{3}{*}{Qwen3-0.6B} & Early & 5.47 & 17.97 & $+12.50\;[+4.41,+21.67]$ \\
         & Middle & 2.34 & 15.63 & $+13.28\;[+6.45,+20.51]$ \\
         & Late & 0.00 & 14.06 & $+14.06\;[+4.61,+25.78]$ \\
        \midrule
        \multirow{3}{*}{Qwen3-4B} & Early & 14.84 & 15.63 & $+0.78\;[-4.84,+6.90]$ \\
         & Middle & 17.97 & 21.09 & $+3.13\;[-9.48,+17.74]$ \\
         & Late & 10.16 & 30.47 & $+20.31\;[+5.47,+37.07]$ \\
        \bottomrule
    \end{tabular}
\end{table}

\paragraph{Paired reverse-KL analysis.}
We run vanilla OPD on ALFWorld for 200 optimizer steps with Qwen3-0.6B and Qwen3-4B students and a frozen Qwen3-32B teacher.
Each step samples 64 student trajectories with temperature $1.0$, top-$p$ $1.0$, and top-$k$ $-1$, with a maximum horizon of 30 turns.
For every sampled response, we evaluate the teacher twice on the same response tokens and student log probabilities, conditioning it once on $h_t$ and once on $s_t$.
We average the sampled reverse-KL estimate first over response tokens within each episode--turn and then over episode--turn values in each optimizer-step--turn-depth cell. At a realized token, the diagnostic uses $\exp(\ell_S-\ell_{\mathrm{proposal}})(\ell_S-\ell_T)$, with the generation-policy log probability in the denominator; unlike the training advantage, this diagnostic is not clamped. It is a sampled estimate, not an exact full-vocabulary KL.
The curves in Figure~\ref{fig:history_intervention}\subref{fig:history_rkl} use a seven-step moving average, whereas the heatmaps use the unsmoothed paired cell means of $\Delta_{\mathrm{RKL}}=\mathrm{RKL}(s_t)-\mathrm{RKL}(h_t)$.
We retain the longest turn prefix for which every optimizer step has at least eight surviving trajectories and at least 10\% of the first-turn trajectory count; both runs satisfy this criterion through turn 30.

\subsection{State-Local Action Diversity}
\label{app:state-coverage-details}

We conduct two separately generated state-level probes from the Qwen3-4B base checkpoint on ALFWorld.
Each probe uses a single pre-update batch of 64 root trajectories, with temperature $1.0$, top-$p$ $1.0$, top-$k$ $-1$, and at most 30 interaction turns.
Each response set contains the realized root response and 15 additional single-turn student responses.
The entropy probe constructs one such set per state, whereas the reasoning-path probe constructs two sets that share the root response.
Branch actions are not executed, and no alternative environment suffix is generated.

\paragraph{Reasoning-path intervention.}
For Figure~\ref{fig:state-coverage}\subref{fig:state-response-space}, we construct two state-matched 16-response sets.
Complete-response resampling regenerates the entire reasoning-to-action response, whereas fixed-reasoning resampling preserves the realized prefix through \texttt{</think>} and regenerates only the action-formatted suffix.
We extract, trim, and lowercase the text between the first \texttt{<action>} and \texttt{</action>} tags, then count distinct nonempty parsed actions within each 16-response set.
Responses without a valid parsed action do not contribute to this count.
We retain 1,615 states with at least two responses containing valid parsed actions under each condition.
The final bar pools sets with five or more distinct actions. Each bar height is the number of states in that action-count group divided by the 1,615 retained states \emph{within its own condition}; the two condition distributions therefore each sum to 100\%.

\paragraph{Entropy--diversity analysis.}
For Figure~\ref{fig:state-coverage}\subref{fig:state-entropy-coverage}, all 15 additional samples regenerate complete responses.
We compute response entropy from the realized root response using Equation~\eqref{eq:response-entropy}.
Additional action diversity is the number of distinct valid parsed actions among the 16 responses, minus one when the realized root response contains a valid action.
The analysis contains 1,831 states.
Markers report means over 10 equal-count entropy bins.
The 95\% confidence intervals use 4,000 nonparametric bootstrap resamples with seed 42, clustered by root trajectory.

\section{Experiment Setup}
\label{app:training-details}

\subsection{Training Hyperparameters}
\label{app:sampled-opd-update}

All runs enable the Qwen3 thinking template. Equation~\eqref{eq:state-opd-objective} gives the ideal full-distribution reverse-KL target for State-OPD; the experiments optimize a sampled-K1 policy surrogate on student-generated tokens. For a sampled token $y_k$, let $\ell_{\theta,k}$ and $\ell_{T,k}$ be its current-student and teacher log probabilities under the respective method's contexts, and let $\ell_{\mathrm{old},k}$ be its generation-policy log probability. We use
\begin{equation}
  A_k=-\operatorname{stopgrad}\!\left(\operatorname{clip}(\ell_{\theta,k}-\ell_{T,k},-10,10)\right),\qquad
  \rho_k=\exp\!\left(\operatorname{clip}(\ell_{\theta,k}-\ell_{\mathrm{old},k},-20,20)\right).
\end{equation}
With $b_k=\max\{-\rho_k A_k,-\operatorname{clip}(\rho_k,0.8,1.2)A_k\}$, the per-token loss is $\min\{b_k,-3A_k\}$ if $A_k<0$ and $b_k$ otherwise. We average this dual-clipped loss over supervised student tokens; environment rewards are not used. Sampled prefixes, teacher scores, cached rollout probabilities, and state selection are treated as fixed during the update. Each configuration has one independent training run with seed 42, so the reported success rates do not estimate variation across training seeds. Table~\ref{tab:training-hyperparameters} lists the fixed training settings, and Table~\ref{tab:method-specific-hyperparameters} gives method-specific parameters.

The 2,048-token capacity applies only to the initial prompt. Subsequent assistant responses and environment observations share a cumulative 32,768-token transcript budget. Responses have no separate per-turn token cap and may use the remaining transcript budget. The accumulated history is retained without sliding-window truncation, and the episode terminates when the remaining budget cannot support further interaction. Observation and appended formatting tokens consume this budget but are excluded from the training loss.

\begin{table}[!htb]
    \centering
    \caption{Training hyperparameters across environments.}
    \label{tab:training-hyperparameters}
    \small
    \begin{tabular}{@{}ll@{}}
        \toprule
        Hyperparameter & Value \\
        \midrule
        \multicolumn{2}{@{}l}{\textbf{Optimization}} \\
        \quad Optimizer & AdamW \\
        \quad Training steps & 200 \\
        \quad Root tasks per step & 16 \\
        \quad Baseline rollouts per task & 4 \\
        \quad Learning rate & $1\times10^{-6}$ (constant) \\
        \quad Gradient clipping & $1.0$ \\
        \quad Policy-update epochs & 1 \\
        \quad Random seed & 42 \\
        \midrule
        \multicolumn{2}{@{}l}{\textbf{Distillation}} \\
        \quad Loss aggregation & token mean \\
        \quad K1 clamp & $[-10,10]$ \\
        \quad Policy-ratio clipping & $0.2$ \\
        \quad Policy log-ratio clamp & $[-20,20]$ \\
        \quad Dual-clip coefficient & $3.0$ \\
        \midrule
        \multicolumn{2}{@{}l}{\textbf{Generation}} \\
        \quad Temperature & $1.0$ \\
        \quad Top-$p$ & $1.0$ \\
        \quad Top-$k$ & $-1$ \\
        \quad Initial-prompt token capacity & 2,048 \\
        \quad Post-initial-prompt transcript token budget & 32,768 \\
        \bottomrule
    \end{tabular}
\end{table}

\subsection{Evaluation Hyperparameters}

All methods use the same prompts and the complete splits in Table~\ref{tab:dataset-statistics}. The context budgets and history retention specified for training in Appendix~\ref{app:sampled-opd-update} also apply to every method at evaluation. Table~\ref{tab:evaluation-hyperparameters} lists the shared decoding settings and environment-specific evaluation limits.

\begin{table}[!htb]
    \centering
    \caption{Evaluation hyperparameters across environments.}
    \label{tab:evaluation-hyperparameters}
    \small
    \begin{tabular}{@{}ll@{}}
        \toprule
        Hyperparameter & Value \\
        \midrule
        \multicolumn{2}{@{}l}{\textbf{Generation}} \\
        \quad Temperature & $0.4$ \\
        \quad Top-$p$ & $1.0$ \\
        \quad Top-$k$ & $-1$ \\
        \midrule
        \multicolumn{2}{@{}l}{\textbf{Evaluation protocol}} \\
        \quad Evaluation interval & 20 training steps \\
        \quad Maximum rounds $H_e$ & 30 (ALFWorld, ScienceWorld); 15 (WebShop) \\
        \quad Samples per task $R_e$ & 4 (ALFWorld, WebShop); 1 (ScienceWorld) \\
        \bottomrule
    \end{tabular}
\end{table}

\section{Benchmark Environments and Metrics}
\label{app:benchmark-details}

\subsection{Benchmark Environments}

We evaluate on three text-based interactive environments spanning embodied planning, scientific reasoning, and web navigation.

\paragraph{ALFWorld.}
ALFWorld~\citep{shridhar2020alfworld} requires an embodied agent to navigate household scenes and manipulate objects through natural-language commands.
Its six task families cover object placement, placement after cleaning, heating, or cooling, examination under a light source, and placement of two objects of the same type.
We retain the official seen and unseen validation splits, which evaluate tasks in training environments and held-out environments, respectively.

\paragraph{ScienceWorld.}
ScienceWorld~\citep{wang2022scienceworld} evaluates interactive scientific reasoning across 30 task types derived from an elementary-school science curriculum.
An agent must carry out multi-step procedures such as locating materials, operating devices, and observing the consequences of interventions.
We report results separately for task types represented during training and held-out task types.

\paragraph{WebShop.}
WebShop~\citep{yao2022webshop} presents a natural-language shopping request and a simulated e-commerce website.
The agent searches the catalogue, navigates product pages and options, and completes a purchase that satisfies the requested attributes.
Evaluation uses held-out shopping sessions.

\subsection{Dataset Sizes and Split Construction}

Table~\ref{tab:dataset-statistics} summarizes the task instances used for distillation and evaluation.
For ALFWorld, we use the original training split and the official in-distribution and out-of-distribution validation splits, yielding 3,553 training games, 140 seen games, and 134 unseen games.
For ScienceWorld, we follow the task-type and variation construction of TCOD~\citep{wang2026tcod}.
Seventeen task types are assigned to the seen setting: the first 50\% of their variations are used for training and the next 10\% for evaluation.
For the remaining 13 held-out task types, the first 30\% of variations form the unseen evaluation set.
We shuffle the resulting splits with a global seed of 42, producing 2,294 training, 456 seen-evaluation, and 782 unseen-evaluation instances.
For WebShop, we use the sequential split of TCOD: session IDs $[0,4096)$ form the training set and IDs $[4096,4596)$ form the 500-session test set.

\begin{table}[!htbp]
    \centering
    \caption{Training and evaluation sets. Parentheses give the number of ScienceWorld task types.}
    \label{tab:dataset-statistics}
    \small
    \begin{tabular}{lrrr}
        \toprule
        Environment & Train & Seen evaluation & Unseen evaluation / test \\
        \midrule
        ALFWorld     & 3,553 & 140 & 134 \\
        ScienceWorld & 2,294 (17) & 456 (17) & 782 (13) \\
        WebShop      & 4,096 & -- & 500 \\
        \bottomrule
    \end{tabular}
\end{table}

\subsection{Metrics}

\paragraph{Success rate (SR $\uparrow$).}
An episode succeeds when ALFWorld reports a win, ScienceWorld reports task completion, or WebShop returns a final reward of 1.
We average binary outcomes across sampled trajectories for each task and then across tasks within each evaluation split.
ALFWorld and WebShop use four trajectories per task, so SR is their mean success rate rather than pass@4; ScienceWorld uses one.

\paragraph{Interaction rounds (Rounds $\downarrow$).}
For successful episodes, we count executed environment actions.
We assign unsuccessful episodes the maximum horizon: 30 rounds in ALFWorld and ScienceWorld, and 15 in WebShop.
We average these values using the same task weighting as SR.
Rounds is therefore a failure-penalized measure rather than the mean number of actions actually executed.

\section{Baseline Implementations}
\label{app:baseline-implementations}

We compare State-OPD with four baselines spanning standard multi-turn OPD, temporal curricula, teacher-guided rollouts, and turn-aware training.
Within each student size, all methods start from the same Qwen3 checkpoint and share the frozen Qwen3-32B teacher, training task pool, optimizer settings, and evaluation protocol described in Appendix~\ref{app:training-details}.
We describe each baseline's mechanism and method-specific settings below; Table~\ref{tab:method-specific-hyperparameters} summarizes the numerical hyperparameters.

\paragraph{Vanilla OPD.}
Vanilla OPD extends on-policy distillation~\citep{agarwal2024policy} to complete multi-turn agent trajectories.
The student generates every reasoning--action response until task termination or the environment horizon, while the teacher scores its response tokens conditioned on the full student-generated history.
We apply the shared clipped sampled-K1 update to these tokens; Vanilla OPD introduces no additional method-specific hyperparameters.

\paragraph{TCOD-F2B.}
TCOD-F2B~\citep{wang2026tcod} introduces a temporal curriculum: the student first learns from short trajectories, then progressively extends its rollout horizon.
We use its recommended starting horizon of one turn and pacing factor $\eta=2$, with the teacher conditioned on the full history.
At optimizer step $k\in\{1,\ldots,K\}$, the student rollout horizon is
\begin{equation}
    H_k = \min\!\left(1+\left\lfloor\frac{k}{\eta}\right\rfloor,H\right),
    \qquad \eta=2,
    \label{eq:tcod-f2b-schedule}
\end{equation}
where $H$ is the full environment horizon.
The curriculum changes the training horizon only; evaluation uses the full environment horizon $H$.

\paragraph{Guided-OPD.}
Guided-OPD~\citep{li2026guidedopd} lets the teacher generate complete reasoning--action turns during student rollouts, then gradually withdraws this intervention.
We follow its recommended cosine curriculum with initial and final teacher-turn probabilities of one and zero and curriculum ratio $\rho=0.8$.
For $K=200$ optimizer steps, the teacher-turn probability at step $k$ is
\begin{equation}
    \beta_k
    = \frac{1}{2}\left[1+\cos\!\left(
        \pi\min\!\left(\frac{k-1}{\rho K},1\right)
    \right)\right].
    \label{eq:guided-opd-schedule}
\end{equation}
The intervention probability reaches zero after the first 160 steps.
We condition the teacher on the full interaction history and train teacher-generated turns with token-level cross-entropy and student-generated turns with the shared clipped sampled-K1 update.
Evaluation uses the student policy alone.

\paragraph{TurnOPD.}
TurnOPD~\citep{zhou2026turnopd} combines an adaptive rollout-depth controller with a progressive shift from trajectory-level to turn-level loss normalization.
We retain both controllers: full-horizon probes estimate a survivor-weighted per-turn sampled-K1 signal and the successful-completion coverage horizon, while other updates use the exponentially smoothed rollout cap.
We use the controller settings reported by the original paper: a minimum depth of two, three warmup probes, a probe every eight steps, EMA coefficient $0.3$, and coverage quantile $0.8$; the remaining controller settings are collected in Table~\ref{tab:method-specific-hyperparameters}.

\begin{table}[!htbp]
    \centering
    \caption{Method-specific hyperparameters.}
    \label{tab:method-specific-hyperparameters}
    \small
    \begin{tabular}{@{}ll@{}}
        \toprule
        Hyperparameter & Value \\
        \midrule
        \multicolumn{2}{@{}l}{\textbf{State-OPD (ours)}} \\
        \quad Concentration factor $\beta$ & 2 \\
        \midrule
        \multicolumn{2}{@{}l}{\textbf{TCOD-F2B}} \\
        \quad Starting horizon & 1 turn \\
        \quad Pacing factor $\eta$ & 2 \\
        \midrule
        \multicolumn{2}{@{}l}{\textbf{Guided-OPD}} \\
        \quad Teacher-turn probability (start / end) & $1 / 0$ \\
        \quad Curriculum ratio $\rho$ & 0.8 \\
        \quad Decay schedule & Cosine \\
        \midrule
        \multicolumn{2}{@{}l}{\textbf{TurnOPD}} \\
        \quad Minimum rollout depth & 2 \\
        \quad Probe warmup / interval & $3 / 8$ steps \\
        \quad Horizon EMA / coverage quantile & $0.3 / 0.8$ \\
        \quad Minimum successful probes & 8 \\
        \quad Loss-blend progress interval & $[0,1]$ \\
        \quad Survivor floor / fraction & $8 / 0.15$ \\
        \bottomrule
    \end{tabular}
\end{table}

\paragraph{Reference policies.}
We evaluate each pre-distillation student checkpoint and the frozen Qwen3-32B teacher without parameter updates.
Both references use the same prompts, environment horizons, and evaluation decoding configuration as the trained policies.

\section{Additional Studies}
\label{app:additional-studies}

\subsection{Paired Bootstrap Uncertainty of Main Results}
\label{app:paired-bootstrap}

Table~\ref{tab:paired-bootstrap-sr} reports success-rate differences between State-OPD and each of the four trained baselines in Table~\ref{tab:main-results}.
We use the step-200 checkpoints from the single training seed (42) and analyze each student size and evaluation setting separately.
The underlying evaluation records jointly reproduce the SR and Rounds entries in the main table to its displayed precision.
Differences and intervals are computed from unrounded evaluation outcomes; rounding is applied only when displaying the results.

\paragraph{Task matching and resampling unit.}
We match methods by stable task identity: the game file in ALFWorld, the task name, variation ID, and simplification setting in ScienceWorld, and the session ID in WebShop.
All five methods have the same complete task set within each comparison: 140/134 Seen/Unseen tasks in ALFWorld, 456/782 in ScienceWorld, and 500 test sessions in WebShop.
For ALFWorld and WebShop, the four evaluations of a task form one resampling unit and are always retained together.
ScienceWorld has one evaluation per task.
Matching task identities does not assume common random-number streams; initial observations can differ across methods in ScienceWorld.

\paragraph{Paired difference intervals.}
Let $y_{i,m,r}\in\{0,1\}$ denote success for task $i$, method $m$, and evaluation repeat $r$, and let $\bar{y}_{i,m}=R_e^{-1}\sum_{r=1}^{R_e}y_{i,m,r}$.
For a baseline $b$, the estimated SR difference in percentage points is
\begin{equation}
    \widehat{\Delta}_{b}
    = \frac{100}{N}\sum_{i=1}^{N}
      \bigl(\bar{y}_{i,\mathrm{State\text{-}OPD}}-\bar{y}_{i,b}\bigr).
    \label{eq:paired-sr-difference}
\end{equation}
For each of 10,000 bootstrap resamples, we draw $N$ task indices with replacement and use the same indices for all five methods.
We recompute the differences on the selected task bundles without resampling evaluations within a task.
The 2.5th and 97.5th percentiles of each difference distribution define its 95\% interval.
Random-number seeds are derived deterministically from a master seed of 2027 for each student--setting pair.
Seen and Unseen are analyzed separately; no interval pools evaluation settings.

\paragraph{Scope of uncertainty.}
These intervals quantify task-level evaluation uncertainty conditional on the fixed trained checkpoints and the observed bundles of evaluation outcomes.
They do not estimate variation across independent training seeds or replace a multiple-seed training study.
The 40 intervals are pointwise 95\% intervals without a multiple-comparison correction, and therefore do not provide simultaneous 95\% coverage of all comparisons.

\begin{table}[!htbp]
    \centering
    \caption{Paired success-rate differences, State-OPD minus the indicated baseline, in percentage points. Brackets give 95\% task-bootstrap intervals. Positive differences favor State-OPD. Each student column contains 20 comparisons.}
    \label{tab:paired-bootstrap-sr}
    \small
    \setlength{\tabcolsep}{4pt}
    \renewcommand{\arraystretch}{1.1}
    \begin{tabular}{@{}llcc@{}}
        \toprule
        Setting & Baseline & Qwen3-1.7B & Qwen3-4B \\
        \midrule
        \multirow{4}{*}{\shortstack[l]{ALFWorld\\Seen}} & Vanilla OPD & $+4.82\;[+0.54,+9.29]$ & $+12.50\;[+8.39,+16.79]$ \\
         & TCOD-F2B & $+2.14\;[-2.32,+6.61]$ & $+14.11\;[+9.64,+18.57]$ \\
         & Guided-OPD & $+3.39\;[-0.89,+7.68]$ & $+16.25\;[+11.43,+21.07]$ \\
         & TurnOPD & $+3.93\;[-0.36,+8.39]$ & $+16.61\;[+11.79,+21.61]$ \\
        \midrule
        \multirow{4}{*}{\shortstack[l]{ALFWorld\\Unseen}} & Vanilla OPD & $+3.54\;[-1.49,+8.58]$ & $+11.75\;[+6.53,+17.35]$ \\
         & TCOD-F2B & $+7.46\;[+2.05,+13.06]$ & $+10.45\;[+5.04,+15.86]$ \\
         & Guided-OPD & $+0.56\;[-4.10,+5.22]$ & $+14.93\;[+9.70,+20.34]$ \\
         & TurnOPD & $-1.68\;[-7.09,+3.92]$ & $+12.13\;[+6.72,+17.72]$ \\
        \midrule
        \multirow{4}{*}{\shortstack[l]{ScienceWorld\\Seen}} & Vanilla OPD & $+0.66\;[-2.41,+3.95]$ & $+1.54\;[-3.29,+6.14]$ \\
         & TCOD-F2B & $+1.54\;[-1.97,+5.04]$ & $+1.32\;[-3.07,+5.92]$ \\
         & Guided-OPD & $-2.19\;[-5.70,+1.32]$ & $-0.44\;[-5.04,+4.39]$ \\
         & TurnOPD & $+2.85\;[-0.22,+6.14]$ & $+3.07\;[-1.32,+7.46]$ \\
        \midrule
        \multirow{4}{*}{\shortstack[l]{ScienceWorld\\Unseen}} & Vanilla OPD & $+0.00\;[-2.81,+2.94]$ & $-0.51\;[-3.32,+2.30]$ \\
         & TCOD-F2B & $+2.05\;[-0.64,+4.73]$ & $+0.26\;[-2.56,+3.20]$ \\
         & Guided-OPD & $-0.13\;[-2.81,+2.69]$ & $-1.15\;[-3.84,+1.41]$ \\
         & TurnOPD & $+2.69\;[+0.13,+5.24]$ & $-0.26\;[-3.07,+2.56]$ \\
        \midrule
        \multirow{4}{*}{\shortstack[l]{WebShop\\Test}} & Vanilla OPD & $+0.30\;[-0.85,+1.50]$ & $+3.10\;[+1.80,+4.50]$ \\
         & TCOD-F2B & $+0.80\;[-0.45,+2.05]$ & $+4.25\;[+2.85,+5.75]$ \\
         & Guided-OPD & $+0.05\;[-1.15,+1.25]$ & $+3.10\;[+1.65,+4.60]$ \\
         & TurnOPD & $-0.60\;[-1.90,+0.70]$ & $+1.85\;[+0.45,+3.30]$ \\
        \bottomrule
    \end{tabular}
\end{table}
\FloatBarrier

\subsection{Failed-Action Repetition Across Training}
\label{app:failed-action-repetition}

\begin{figure}[t]
    \centering
    \includegraphics[width=\linewidth]{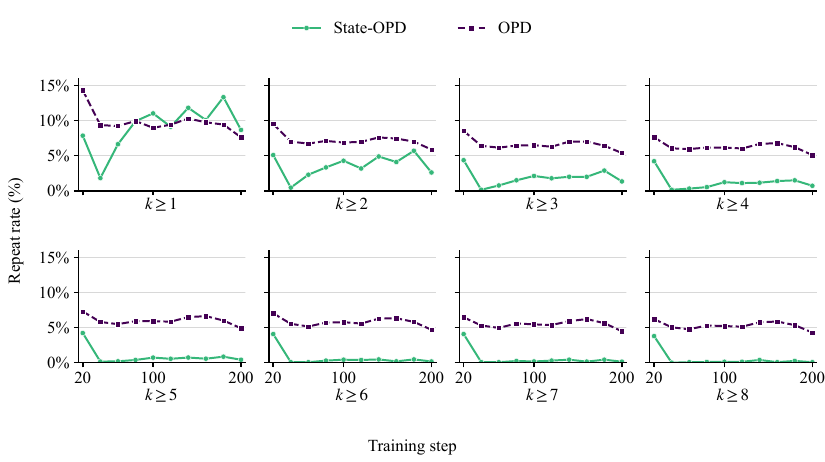}
    \caption{State-OPD reduces persistent failed-action loops throughout ALFWorld training. For a Qwen3-4B student distilled from Qwen3-32B, each panel shows the share of failed-action starts with at least $k$ further identical actions, each returning \texttt{Nothing happens.}}
    \label{fig:failed-action-repeat-training}
\end{figure}

Figure~\ref{fig:failed-action-repeat-training} expands the step-200 comparison in Figure~\ref{fig:failed-action-repeat-depth} to all ten evaluation checkpoints. The curves separate much more strongly at deeper repeat depths than at the first retry: at $k=5$, State-OPD remains below 0.9\% at every checkpoint from step 40 onward, whereas Vanilla OPD remains above 4.9\%.

\subsection{State Allocation Analysis and Sensitivity}
\label{app:allocation-analysis}

The concentration factor $\beta$ sets the breadth--depth trade-off of response allocation under a nominally matched turn-response budget: larger values sample more responses at fewer interaction states. For a root trajectory of length $T_i$ and reference multiplicity $M=4$, State-OPD selects $P_i=\lceil T_i/\beta\rceil$ states and samples $n=\beta(M-1)$ additional responses at each selected state. We use $\beta=2$ across environments and model sizes, assigning six additional responses to approximately half of the states in each trajectory.

\begin{figure*}[t]
    \newlength{\stateallocationpanelheight}
    \setlength{\stateallocationpanelheight}{0.40\textwidth}
    \centering
    \begin{subfigure}[t]{0.47\textwidth}
        \centering
        \vspace{0pt}
        \begin{minipage}[c][\stateallocationpanelheight][c]{\linewidth}
            \centering
            \includegraphics[width=\linewidth]{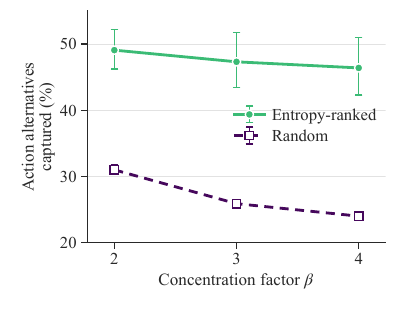}
        \end{minipage}
        \caption{Observed action diversity captured}
        \label{fig:beta-matched-coverage}
    \end{subfigure}
    \hfill
    \begin{subfigure}[t]{0.49\textwidth}
        \centering
        \vspace{0pt}
        \begin{minipage}[c][\stateallocationpanelheight][c]{\linewidth}
            \centering
            \scriptsize
            \setlength{\tabcolsep}{3.2pt}
            \renewcommand{\arraystretch}{1.12}
            \begin{tabular}{lcc}
                \toprule
                Configuration & Seen SR $\uparrow$ & Unseen SR $\uparrow$ \\
                \midrule
                Vanilla OPD & 30.89 & 33.96 \\
                \midrule
                State-OPD, $\beta=2$ (main) & 43.39 & 45.71 \\
                State-OPD, $\beta=3$ & 44.46 & 46.83 \\
                State-OPD, $\beta=4$ & 40.18 & 44.59 \\
                \bottomrule
            \end{tabular}
        \end{minipage}
        \caption{Success-rate sensitivity to $\beta$}
        \label{fig:allocation-concentration-sensitivity}
    \end{subfigure}
    \caption{State selection and allocation sensitivity on ALFWorld. (a) Share of additional distinct actions in a 16-response probe recovered by entropy-ranked or random selection under matched budgets; error bars are 95\% root-cluster bootstrap intervals. (b) Seen/Unseen SR across $\beta$ for a Qwen3-4B student distilled from Qwen3-32B; each setting has one training run.}
    \label{fig:allocation-analysis}
\end{figure*}

\paragraph{Action diversity at matched budgets.}
We ask whether concentrating samples on high-entropy states exposes more actions under the same budget. A pre-update ALFWorld probe contains one root response and 15 additional complete responses at every visited state. For each tested $\beta\in\{2,3,4\}$, entropy ranking and random selection choose the same number $P_i$ of states within each root trajectory; at each selected state, we analytically average over uniform choices of $n$ responses from the 15 available. We measure the additional distinct parsed actions beyond the root action as a fraction of those observed in the complete probe. At $\beta=2$, Figure~\ref{fig:beta-matched-coverage} shows that entropy ranking captures 49.09\% of the observed action alternatives, versus 30.99\% for random selection. Entropy ranking also captures a larger fraction of the observed action diversity at $\beta=3$ and $4$, while $\beta=2$ gives its highest captured fraction. This provides an action-diversity rationale for the common default.

\paragraph{Sensitivity to the concentration factor.}
We next vary $\beta\in\{2,3,4\}$ in ALFWorld training with a Qwen3-4B student and Qwen3-32B teacher. As Figure~\ref{fig:allocation-concentration-sensitivity} shows, Seen success ranges from 40.18\% to 44.46\% across these settings, compared with 30.89\% for Vanilla OPD; Unseen success ranges from 44.59\% to 46.83\%, compared with 33.96\%. The highest observed scores on both splits occur at $\beta=3$, while performance declines at $\beta=4$. Thus, the gain over OPD persists across the tested breadth--depth allocations.

\section{Environment Prompts}
\label{app:environment-prompts}

State-OPD changes the trajectory context supplied to the teacher, as described in Section~\ref{sec:state-opd}, while keeping the system instruction and environment interface fixed across methods and between training and evaluation.
The following listings reproduce the system prompts used by the implementation.
The task and initial observation are supplied in the first user message, and subsequent environment observations are supplied as tool messages.

\lstdefinestyle{environmentprompt}{
    basicstyle=\ttfamily\scriptsize,
    breaklines=true,
    breakatwhitespace=true,
    columns=fullflexible,
    keepspaces=true,
    showstringspaces=false,
    frame=single,
    aboveskip=0.5em,
    belowskip=0.5em
}

\subsection{ALFWorld}
\begin{lstlisting}[style=environmentprompt]
You are an agent in the ALFRED embodied household environment. The first user message gives the initial scene and task. After each action, a tool message gives the environment's new observation.

## Available commands
  look:                             look around your current location
  inventory:                        check your current inventory
  go to (receptacle):               move to a receptacle
  open (receptacle):                open a receptacle
  close (receptacle):               close a receptacle
  take (object) from (receptacle):  take an object from a receptacle
  move (object) to (receptacle):    place an object in or on a receptacle
  examine (something):              examine a receptacle or an object
  use (object):                     use an object
  heat (object) with (receptacle):  heat an object using a receptacle
  clean (object) with (receptacle): clean an object using a receptacle
  cool (object) with (receptacle):  cool an object using a receptacle
  slice (object) with (object):     slice an object using a sharp object

## Response format
First reason about the current situation inside <think> </think> tags. Then output exactly one action inside <action> </action> tags.
If the environment replies "Nothing happens.", the previous action was invalid or used the wrong command format; try a different valid action.
\end{lstlisting}

\subsection{ScienceWorld}
\begin{lstlisting}[style=environmentprompt]
You are an agent in the ScienceWorld text environment.
The first user message gives the task and initial scene.
After each action, a tool message gives the environment's new observation.

## Available commands
  open OBJ: open a door or container
  close OBJ: close a door or container
  activate OBJ: activate a device
  deactivate OBJ: deactivate a device
  use OBJ on OBJ: use a device on an object
  dunk OBJ in OBJ: dunk a container into a liquid
  eat OBJ: eat an edible object
  flush OBJ: flush a toilet
  inventory: list the objects in your inventory
  look around: describe the current room
  look at OBJ: inspect an object
  look in OBJ: inspect a container
  read OBJ: read an object
  go to LOC: move to an adjacent location
  move OBJ to OBJ: move an object into or onto another object
  pick up OBJ: move an object to the inventory
  put down OBJ: put down an inventory object
  pour OBJ in OBJ: pour one container into another
  mix OBJ: mix the contents of a container
  teleport to LOC: move directly to a room
  focus on OBJ: select an object as the task focus
  task: repeat the task description
  wait: wait for 10 simulator iterations
  wait1: wait for 1 simulator iteration

## Response format
First reason about the current situation inside <think> </think> tags.
Then output exactly one action inside <action> </action> tags.
\end{lstlisting}

For task names containing \texttt{power-component} or \texttt{conductivity}, the following two command lines are appended to the available-command list:
\begin{lstlisting}[style=environmentprompt]
  connect OBJ to OBJ: connect two electrical terminals
  disconnect OBJ: disconnect an electrical component
\end{lstlisting}

\subsection{WebShop}
\begin{lstlisting}[style=environmentprompt]
You are an agent interacting with a virtual text-based web shopping environment to test out your ability. Your job is to follow the Instruction provided and mimic the steps to buy the item that are closest to the Instruct provided.
The first user message gives the shopping task and initial page.
After each action, a tool message gives the environment's new observation.

## Available commands
You should give both the action_name and action_arg like the format `action_name[action_arg]`. You can execute two types of actions, search and click.
  search[QUERY]: search for products only when the button `[button] Search [button_]` is available in the current observation
  click[BUTTON]: click buttons `[button] xxx [button_]` that is available in the current observation

## Notes:
At each step, you should first think then perform action to fulfill the instruction. You should ALWAYS wrap your thinking with the <think> </think> tag and wrap your action with the <action> </action> tag.
You should ALWAYS take one action each step.
You should finish the task and buy the item within {MAX_EPISODE_STEPS} steps.
DO NOT try to interact with the user at anytime. Finish the task and buy the item by yourself.
\end{lstlisting}

\raggedbottom
\section{Case Studies}
\label{app:case-studies}
Each case pairs step-200 validation rollouts of State-OPD and Vanilla OPD, both Qwen3-4B students distilled from a Qwen3-32B teacher. These qualitative examples are drawn from separate step-200 diagnostic runs. They illustrate observed reasoning and action patterns rather than identify the cause or magnitude of the main-table improvements. Each pair shares a task and initial input; the panels account for every turn's reason and action, and show the decisive observations and outcomes. In these traces, \textbf{Reason:} \ldots{} denotes omitted reasoning, including for each turn in a grouped sequence.

\definecolor{traceTask}{HTML}{314B80}
\definecolor{traceState}{HTML}{16805F}
\definecolor{traceOpd}{HTML}{60306F}
\newcommand{\tracepanel}[3]{%
  \begin{tcolorbox}[
    breakable,
    sharp corners,
    colframe=#1,
    colback=#1!5,
    colbacktitle=#1,
    coltitle=white,
    fonttitle=\bfseries,
    title={#2},
    boxrule=0.4pt,
    boxsep=5pt,
    left=5pt,
    right=5pt,
    top=3pt,
    bottom=3pt,
    before skip=7pt,
    after skip=9pt]
    \setlength{\parindent}{0pt}%
    \setlength{\parskip}{2pt}%
    #3
  \end{tcolorbox}%
}

\subsection{Re-examining the Interaction State Before Acting}
\label{app:state-review-case}
To illustrate how students revisit interaction history, we examine task-matched evaluation traces that share the same $s_t$ at the first divergent action. These examples show the State-OPD student reviewing task requirements and prior observations before choosing its next action. We reproduce both students' complete thinking blocks at that point. Each turn lists its reason, action, and observation; only identical action--observation pairs repeated for more than five consecutive turns are grouped.
The pan pair uses ALFWorld game \texttt{trial\_T20190908\_032543\_712058} and validation repeats 0 (State-OPD) and 1 (Vanilla OPD); the mug pair uses \texttt{trial\_T20190907\_221208\_560499} and repeats 2 and 1, respectively. The source index \texttt{state\_review\_case\_sources.json} records both pairs' compact-log locations and initial-prompt hashes.

\paragraph{Reconsidering where to search.}
After 15 identical action--observation pairs, both students have searched the same cabinets and drawers without finding the pan. State-OPD reviews the observed locations, chooses countertop 1, then takes the pan, cleans it at the sinkbasin, and places it on the countertop. Vanilla OPD searches shelves and the fridge before finding the pan at stoveburner 1; after taking it, the student attempts placement and cleaning without navigating to the required receptacles and reaches the turn limit. In the game's initial PDDL, the same concrete pan object has \texttt{inReceptacle} relations with two stove burners and one countertop, accounting for the different locations reported by these traces from the shared game state. The case illustrates the students' action choices after a shared history and their execution of the cleaning-and-placement sequence. Across four evaluations of this task, State-OPD succeeds in 4/4 and Vanilla OPD in 0/4.

\begingroup
\footnotesize
\tracepanel{traceTask}{Task and shared actions through turn 15}{%
  \textbf{Task.} \texttt{put a clean pan in countertop.}\par\vspace{1pt}
  \textbf{Obs 0:} ``-= Welcome to TextWorld, ALFRED! =-''\newline
  ``You are in the middle of a room. Looking quickly around you, you see a cabinet 6, a cabinet 5, a cabinet 4, a cabinet 3, a cabinet 2, a cabinet 1, a coffeemachine 1, a countertop 3, a countertop 2, a countertop 1, a drawer 3, a drawer 2, a drawer 1, a fridge 1, a garbagecan 1, a microwave 1, a shelf 3, a shelf 2, a shelf 1, a sinkbasin 1, a stoveburner 4, a stoveburner 3, a stoveburner 2, a stoveburner 1, and a toaster 1.''\newline
  ``Your task is to: put a clean pan in countertop.''\par\vspace{1pt}
  \textbf{Turn 1.}\newline
  \textbf{Reason:} \ldots{}\newline
  \textbf{Action:} \texttt{go to cabinet 1}.\newline
  \textbf{Observation:} ``You arrive at cabinet 1. On the cabinet 1, you see a dishsponge 1.''\par\vspace{1pt}
  \textbf{Turn 2.}\newline
  \textbf{Reason:} \ldots{}\newline
  \textbf{Action:} \texttt{go to cabinet 2}.\newline
  \textbf{Observation:} ``You arrive at cabinet 2. The cabinet 2 is closed.''\par\vspace{1pt}
  \textbf{Turn 3.}\newline
  \textbf{Reason:} \ldots{}\newline
  \textbf{Action:} \texttt{open cabinet 2}.\newline
  \textbf{Observation:} ``You open the cabinet 2. The cabinet 2 is open. In it, you see a bowl 1, and a cup 1.''\par\vspace{1pt}
  \textbf{Turn 4.}\newline
  \textbf{Reason:} \ldots{}\newline
  \textbf{Action:} \texttt{go to cabinet 3}.\newline
  \textbf{Observation:} ``You arrive at cabinet 3. On the cabinet 3, you see a glassbottle 1.''\par\vspace{1pt}
  \textbf{Turn 5.}\newline
  \textbf{Reason:} \ldots{}\newline
  \textbf{Action:} \texttt{go to cabinet 4}.\newline
  \textbf{Observation:} ``You arrive at cabinet 4. On the cabinet 4, you see a plate 1.''\par\vspace{1pt}
  \textbf{Turn 6.}\newline
  \textbf{Reason:} \ldots{}\newline
  \textbf{Action:} \texttt{go to cabinet 5}.\newline
  \textbf{Observation:} ``You arrive at cabinet 5. The cabinet 5 is closed.''\par\vspace{1pt}
  \textbf{Turn 7.}\newline
  \textbf{Reason:} \ldots{}\newline
  \textbf{Action:} \texttt{open cabinet 5}.\newline
  \textbf{Observation:} ``You open the cabinet 5. The cabinet 5 is open. In it, you see a peppershaker 2.''\par\vspace{1pt}
  \textbf{Turn 8.}\newline
  \textbf{Reason:} \ldots{}\newline
  \textbf{Action:} \texttt{go to cabinet 6}.\newline
  \textbf{Observation:} ``You arrive at cabinet 6. The cabinet 6 is closed.''\par\vspace{1pt}
  \textbf{Turn 9.}\newline
  \textbf{Reason:} \ldots{}\newline
  \textbf{Action:} \texttt{open cabinet 6}.\newline
  \textbf{Observation:} ``You open the cabinet 6. The cabinet 6 is open. In it, you see a saltshaker 1, and a soapbottle 1.''\par\vspace{1pt}
  \textbf{Turn 10.}\newline
  \textbf{Reason:} \ldots{}\newline
  \textbf{Action:} \texttt{go to drawer 1}.\newline
  \textbf{Observation:} ``You arrive at drawer 1. The drawer 1 is closed.''\par\vspace{1pt}
  \textbf{Turn 11.}\newline
  \textbf{Reason:} \ldots{}\newline
  \textbf{Action:} \texttt{open drawer 1}.\newline
  \textbf{Observation:} ``You open the drawer 1. The drawer 1 is open. In it, you see a fork 1, and a spatula 1.''\par\vspace{1pt}
  \textbf{Turn 12.}\newline
  \textbf{Reason:} \ldots{}\newline
  \textbf{Action:} \texttt{go to drawer 2}.\newline
  \textbf{Observation:} ``You arrive at drawer 2. The drawer 2 is closed.''\par\vspace{1pt}
  \textbf{Turn 13.}\newline
  \textbf{Reason:} \ldots{}\newline
  \textbf{Action:} \texttt{open drawer 2}.\newline
  \textbf{Observation:} ``You open the drawer 2. The drawer 2 is open. In it, you see a creditcard 1, a saltshaker 2, and a spoon 1.''\par\vspace{1pt}
  \textbf{Turn 14.}\newline
  \textbf{Reason:} \ldots{}\newline
  \textbf{Action:} \texttt{go to drawer 3}.\newline
  \textbf{Observation:} ``You arrive at drawer 3. The drawer 3 is closed.''\par\vspace{1pt}
  \textbf{Turn 15.}\newline
  \textbf{Reason:} \ldots{}\newline
  \textbf{Action:} \texttt{open drawer 3}.\newline
  \textbf{Observation:} ``You open the drawer 3. The drawer 3 is open. In it, you see a creditcard 2, and a peppershaker 1.''\par
}
\tracepanel{traceState}{State-OPD \hfill Task completed in 21 turns}{%
  \textbf{Turn 16.}\newline
  \textbf{Reason:} ``Okay, let me try to figure out what to do next. The task is to put a clean pan in the countertop. So far, I've checked several cabinets and drawers but haven't found a pan yet. Let me recap where I've looked:

  \par\noindent - Cabinet 1: dishsponge
  \par\noindent - Cabinet 2: bowl, cup
  \par\noindent - Cabinet 3: glassbottle
  \par\noindent - Cabinet 4: plate
  \par\noindent - Cabinet 5: peppershaker
  \par\noindent - Cabinet 6: saltshaker, soapbottle
  \par\noindent - Drawer 1: fork, spatula
  \par\noindent - Drawer 2: creditcard, saltshaker, spoon
  \par\noindent - Drawer 3: creditcard, peppershaker

  Hmm, no pan found in these places. Maybe the pan is in another location. Let me think about other possible places where a pan might be. The user mentioned a fridge, sinkbasin, stoveburners, etc. Maybe the pan is in the fridge or on the stove. Let me check the fridge first. If the pan is in the fridge, I need to take it out and then clean it if necessary. Wait, the task says "clean pan," so maybe the pan is already clean, or I need to clean it. But first, I need to find the pan.

  Alternatively, maybe the pan is on the countertop already. Let me check the countertops. The user mentioned countertops 1, 2, 3. Let me go to countertop 1 and see if there's a pan there. If not, check the others. If the pan is on a countertop, I need to take it and then put it back, but the task is to put a clean one there. Wait, the task is to put a clean pan in the countertop. So maybe the pan is already in a location, but not on the countertop. So I need to find it, take it, clean it if necessary, and then place it on the countertop.

  Wait, the user hasn't checked the countertops yet. Let me check the countertops. Let's start with countertop 1. If there's a pan there, maybe it's already clean. If not, maybe check the other countertops. If the pan is on a countertop, I can take it and then put it back. But if it's not there, I need to find it elsewhere. Let's check the countertops first.

  So the next step is to go to countertop 1 and look for a pan. If there's a pan there, take it. If not, check the other countertops. If the pan is in another location, like the fridge or on the stove, then I need to find it there. Let me check the countertops first.''\par
  \textbf{Action:} \texttt{go to countertop 1}.\newline
  \textbf{Observation:} ``You arrive at countertop 1. On the countertop 1, you see a mug 2, and a pan 1.''\par\vspace{1pt}
  \textbf{Turn 17.}\newline
  \textbf{Reason:} \ldots{}\newline
  \textbf{Action:} \texttt{take pan 1 from countertop 1}.\newline
  \textbf{Observation:} ``You pick up the pan 1 from the countertop 1.''\par\vspace{1pt}
  \textbf{Turn 18.}\newline
  \textbf{Reason:} \ldots{}\newline
  \textbf{Action:} \texttt{go to sinkbasin 1}.\newline
  \textbf{Observation:} ``You arrive at sinkbasin 1. On the sinkbasin 1, you see nothing.''\par\vspace{1pt}
  \textbf{Turn 19.}\newline
  \textbf{Reason:} \ldots{}\newline
  \textbf{Action:} \texttt{clean pan 1 with sinkbasin 1}.\newline
  \textbf{Observation:} ``You clean the pan 1 using the sinkbasin 1.''\par\vspace{1pt}
  \textbf{Turn 20.}\newline
  \textbf{Reason:} \ldots{}\newline
  \textbf{Action:} \texttt{go to countertop 1}.\newline
  \textbf{Observation:} ``You arrive at countertop 1. On the countertop 1, you see a mug 2.''\par\vspace{1pt}
  \textbf{Turn 21.}\newline
  \textbf{Reason:} \ldots{}\newline
  \textbf{Action:} \texttt{move pan 1 to countertop 1}.\newline
  \textbf{Observation:} ``You move the pan 1 to the countertop 1.''\newline
  \textbf{Outcome:} Task completed.\par
}
\tracepanel{traceOpd}{Vanilla OPD \hfill Failure at the 30-turn limit}{%
  \textbf{Turn 16.}\newline
  \textbf{Reason:} ``Okay, so I need to put a clean pan on the countertop. Let me recap where I've looked so far. I checked all the cabinets (1-6), drawers 1, 2, and 3, and found various items like dishsponge, bowls, cups, glassbottles, plates, peppershakers, saltshakers, and utensils, but no pan. The next step is to check the shelves. There are shelves 1, 2, and 3. Maybe the pan is on one of them. Let me go to shelf 1 and check. If it's not there, I'll move to shelf 2 or 3. Once I find the pan, I need to check if it's clean. If it's dirty, I might need to clean it using the sinkbasin. After that, I can place it on the countertop. Let's start with shelf 1.''\par
  \textbf{Action:} \texttt{go to shelf 1}.\newline
  \textbf{Observation:} ``You arrive at shelf 1. On the shelf 1, you see a statue 1.''\par\vspace{1pt}
  \textbf{Turn 17.}\newline
  \textbf{Reason:} \ldots{}\newline
  \textbf{Action:} \texttt{go to shelf 2}.\newline
  \textbf{Observation:} ``You arrive at shelf 2. On the shelf 2, you see a cup 3, a papertowelroll 1, a peppershaker 3, and a vase 1.''\par\vspace{1pt}
  \textbf{Turn 18.}\newline
  \textbf{Reason:} \ldots{}\newline
  \textbf{Action:} \texttt{go to shelf 3}.\newline
  \textbf{Observation:} ``You arrive at shelf 3. On the shelf 3, you see a vase 2.''\par\vspace{1pt}
  \textbf{Turn 19.}\newline
  \textbf{Reason:} \ldots{}\newline
  \textbf{Action:} \texttt{go to sinkbasin 1}.\newline
  \textbf{Observation:} ``You arrive at sinkbasin 1. On the sinkbasin 1, you see nothing.''\par\vspace{1pt}
  \textbf{Turn 20.}\newline
  \textbf{Reason:} \ldots{}\newline
  \textbf{Action:} \texttt{go to fridge 1}.\newline
  \textbf{Observation:} ``You arrive at fridge 1. The fridge 1 is closed.''\par\vspace{1pt}
  \textbf{Turn 21.}\newline
  \textbf{Reason:} \ldots{}\newline
  \textbf{Action:} \texttt{open fridge 1}.\newline
  \textbf{Observation:} ``You open the fridge 1. The fridge 1 is open. In it, you see a apple 3, a apple 2, a apple 1, a mug 1, a potato 3, a potato 2, and a potato 1.''\par\vspace{1pt}
  \textbf{Turn 22.}\newline
  \textbf{Reason:} \ldots{}\newline
  \textbf{Action:} \texttt{go to stoveburner 1}.\newline
  \textbf{Observation:} ``You arrive at stoveburner 1. On the stoveburner 1, you see a pan 1.''\par\vspace{1pt}
  \textbf{Turn 23.}\newline
  \textbf{Reason:} \ldots{}\newline
  \textbf{Action:} \texttt{take pan 1 from stoveburner 1}.\newline
  \textbf{Observation:} ``You pick up the pan 1 from the stoveburner 1.''\par\vspace{1pt}
  \textbf{Turn 24.}\newline
  \textbf{Reason:} \ldots{}\newline
  \textbf{Action:} \texttt{examine pan 1}.\newline
  \textbf{Observation:} ``This is a normal pan 1. In it, you see nothing.''\par\vspace{1pt}
  \textbf{Turn 25.}\newline
  \textbf{Reason:} \ldots{}\newline
  \textbf{Action:} \texttt{move pan 1 to countertop 1}.\newline
  \textbf{Observation:} \texttt{Nothing happens.}\par\vspace{1pt}
  \textbf{Turn 26.}\newline
  \textbf{Reason:} \ldots{}\newline
  \textbf{Action:} \texttt{move pan 1 to countertop 2}.\newline
  \textbf{Observation:} \texttt{Nothing happens.}\par\vspace{1pt}
  \textbf{Turn 27.}\newline
  \textbf{Reason:} \ldots{}\newline
  \textbf{Action:} \texttt{move pan 1 to countertop 3}.\newline
  \textbf{Observation:} \texttt{Nothing happens.}\par\vspace{1pt}
  \textbf{Turn 28.}\newline
  \textbf{Reason:} \ldots{}\newline
  \textbf{Action:} \texttt{move pan 1 to countertop 1}.\newline
  \textbf{Observation:} \texttt{Nothing happens.}\par\vspace{1pt}
  \textbf{Turn 29.}\newline
  \textbf{Reason:} \ldots{}\newline
  \textbf{Action:} \texttt{move pan 1 to countertop 1}.\newline
  \textbf{Observation:} \texttt{Nothing happens.}\par\vspace{1pt}
  \textbf{Turn 30.}\newline
  \textbf{Reason:} \ldots{}\newline
  \textbf{Action:} \texttt{clean pan 1 with sinkbasin 1}.\newline
  \textbf{Observation:} \texttt{Nothing happens.}\newline
  \textbf{Outcome:} Turn limit reached.\par
}
\endgroup

\paragraph{Cleaning a mug before placing it.}
After ten identical action--observation pairs, both students hold the same mug. State-OPD checks its current location and the outstanding cleaning requirement, then carries out navigation, cleaning, and placement in order. Vanilla OPD also acknowledges the requirement but subsequently places the mug without cleaning it. The contrast concerns whether the stated prerequisite governs the executed plan. Across four evaluations of this task, State-OPD succeeds in 4/4 and Vanilla OPD in 1/4.

\begingroup
\footnotesize
\tracepanel{traceTask}{Task and shared actions through turn 10}{%
  \textbf{Task.} \texttt{put a clean mug in coffeemachine.}\par\vspace{1pt}
  \textbf{Obs 0:} ``-= Welcome to TextWorld, ALFRED! =-''\newline
  ``You are in the middle of a room. Looking quickly around you, you see a cabinet 6, a cabinet 5, a cabinet 4, a cabinet 3, a cabinet 2, a cabinet 1, a coffeemachine 1, a countertop 3, a countertop 2, a countertop 1, a drawer 3, a drawer 2, a drawer 1, a fridge 1, a garbagecan 1, a microwave 1, a shelf 3, a shelf 2, a shelf 1, a sinkbasin 1, a stoveburner 4, a stoveburner 3, a stoveburner 2, a stoveburner 1, and a toaster 1.''\newline
  ``Your task is to: put a clean mug in coffeemachine.''\par\vspace{1pt}
  \textbf{Turn 1.}\newline
  \textbf{Reason:} \ldots{}\newline
  \textbf{Action:} \texttt{go to cabinet 1}.\newline
  \textbf{Observation:} ``You arrive at cabinet 1. On the cabinet 1, you see a vase 2.''\par\vspace{1pt}
  \textbf{Turn 2.}\newline
  \textbf{Reason:} \ldots{}\newline
  \textbf{Action:} \texttt{go to cabinet 2}.\newline
  \textbf{Observation:} ``You arrive at cabinet 2. The cabinet 2 is closed.''\par\vspace{1pt}
  \textbf{Turn 3.}\newline
  \textbf{Reason:} \ldots{}\newline
  \textbf{Action:} \texttt{open cabinet 2}.\newline
  \textbf{Observation:} ``You open the cabinet 2. The cabinet 2 is open. In it, you see a dishsponge 2, and a vase 3.''\par\vspace{1pt}
  \textbf{Turn 4.}\newline
  \textbf{Reason:} \ldots{}\newline
  \textbf{Action:} \texttt{go to cabinet 3}.\newline
  \textbf{Observation:} ``You arrive at cabinet 3. On the cabinet 3, you see a bowl 1.''\par\vspace{1pt}
  \textbf{Turn 5.}\newline
  \textbf{Reason:} \ldots{}\newline
  \textbf{Action:} \texttt{go to cabinet 4}.\newline
  \textbf{Observation:} ``You arrive at cabinet 4. On the cabinet 4, you see a vase 1.''\par\vspace{1pt}
  \textbf{Turn 6.}\newline
  \textbf{Reason:} \ldots{}\newline
  \textbf{Action:} \texttt{go to cabinet 5}.\newline
  \textbf{Observation:} ``You arrive at cabinet 5. The cabinet 5 is closed.''\par\vspace{1pt}
  \textbf{Turn 7.}\newline
  \textbf{Reason:} \ldots{}\newline
  \textbf{Action:} \texttt{open cabinet 5}.\newline
  \textbf{Observation:} ``You open the cabinet 5. The cabinet 5 is open. In it, you see a soapbottle 1.''\par\vspace{1pt}
  \textbf{Turn 8.}\newline
  \textbf{Reason:} \ldots{}\newline
  \textbf{Action:} \texttt{go to cabinet 6}.\newline
  \textbf{Observation:} ``You arrive at cabinet 6. The cabinet 6 is closed.''\par\vspace{1pt}
  \textbf{Turn 9.}\newline
  \textbf{Reason:} \ldots{}\newline
  \textbf{Action:} \texttt{open cabinet 6}.\newline
  \textbf{Observation:} ``You open the cabinet 6. The cabinet 6 is open. In it, you see a cup 1, and a mug 3.''\par\vspace{1pt}
  \textbf{Turn 10.}\newline
  \textbf{Reason:} \ldots{}\newline
  \textbf{Action:} \texttt{take mug 3 from cabinet 6}.\newline
  \textbf{Observation:} ``You pick up the mug 3 from the cabinet 6.''\par
}
\tracepanel{traceState}{State-OPD \hfill Task completed in 14 turns}{%
  \textbf{Turn 11.}\newline
  \textbf{Reason:} ``Okay, so I need to put a clean mug in the coffeemachine. Let me think. I just took the mug 3 from cabinet 6. Now, I need to make sure it's clean. If it's not clean, I have to clean it. How do I clean an object? The available commands include clean (object) with (receptacle). So maybe I can use the sinkbasin 1 to clean the mug. Let me check if the sink is available. The environment mentions a sinkbasin 1. So I should go to the sinkbasin, then clean the mug there.

  Wait, but first, I need to go to the sinkbasin. Let me see. The current location is cabinet 6. So I need to go to the sinkbasin 1. Then use the clean command. Once it's clean, I can put it into the coffeemachine. But wait, the coffeemachine is already in the initial list. So after cleaning, I can take the mug to the coffeemachine and put it in. Let me proceed step by step.

  First, go to the sinkbasin 1. Then clean the mug 3 with the sinkbasin. Then go to the coffeemachine 1 and put the mug there. Let me check if the coffeemachine is open. If it's closed, I need to open it first. But the initial observation didn't mention the coffeemachine being closed. Hmm. Maybe I need to open it. Wait, the task is to put the mug into the coffeemachine. So maybe the coffeemachine is a receptacle that can be opened. Let me check.

  But first, let me go to the sinkbasin. Let's try that. So next action: go to sinkbasin 1. Then clean the mug. Then go to coffeemachine 1, open it if necessary, and put the mug in.''\par
  \textbf{Action:} \texttt{go to sinkbasin 1}.\newline
  \textbf{Observation:} ``You arrive at sinkbasin 1. On the sinkbasin 1, you see a plate 3, a spatula 2, and a spatula 1.''\par\vspace{1pt}
  \textbf{Turn 12.}\newline
  \textbf{Reason:} \ldots{}\newline
  \textbf{Action:} \texttt{clean mug 3 with sinkbasin 1}.\newline
  \textbf{Observation:} ``You clean the mug 3 using the sinkbasin 1.''\par\vspace{1pt}
  \textbf{Turn 13.}\newline
  \textbf{Reason:} \ldots{}\newline
  \textbf{Action:} \texttt{go to coffeemachine 1}.\newline
  \textbf{Observation:} ``You arrive at coffeemachine 1. On the coffeemachine 1, you see nothing.''\par\vspace{1pt}
  \textbf{Turn 14.}\newline
  \textbf{Reason:} \ldots{}\newline
  \textbf{Action:} \texttt{move mug 3 to coffeemachine 1}.\newline
  \textbf{Observation:} ``You move the mug 3 to the coffeemachine 1.''\newline
  \textbf{Outcome:} Task completed.\par
}
\tracepanel{traceOpd}{Vanilla OPD \hfill Failure at the 30-turn limit}{%
  \textbf{Turn 11.}\newline
  \textbf{Reason:} ``Okay, so I've taken the mug 3 from cabinet 6. Now I need to check if it's clean. The task says "put a clean mug in coffeemachine," so if the mug isn't clean, I have to clean it first. How do I check if it's clean? Maybe I can examine it. Let me use the examine command on the mug 3. If it's dirty, I'll need to clean it using the sinkbasin. Once it's clean, I can put it in the coffeemachine. Let's examine the mug first.''\par
  \textbf{Action:} \texttt{examine mug 3}.\newline
  \textbf{Observation:} ``This is a normal mug 3. In it, you see nothing.''\par\vspace{1pt}
  \textbf{Turn 12.}\newline
  \textbf{Reason:} \ldots{}\newline
  \textbf{Action:} \texttt{go to coffeemachine 1}.\newline
  \textbf{Observation:} ``You arrive at coffeemachine 1. On the coffeemachine 1, you see nothing.''\par\vspace{1pt}
  \textbf{Turn 13.}\newline
  \textbf{Reason:} \ldots{}\newline
  \textbf{Action:} \texttt{open coffeemachine 1}.\newline
  \textbf{Observation:} \texttt{Nothing happens.}\par\vspace{1pt}
  \textbf{Turn 14.}\newline
  \textbf{Reason:} \ldots{}\newline
  \textbf{Action:} \texttt{move mug 3 to coffeemachine 1}.\newline
  \textbf{Observation:} ``You move the mug 3 to the coffeemachine 1.''\par\vspace{1pt}
  \textbf{Turn 15.}\newline
  \textbf{Reason:} \ldots{}\newline
  \textbf{Action:} \texttt{[empty]}.\newline
  \textbf{Observation:} \texttt{Nothing happens.}\par\vspace{1pt}
  \textbf{Turn 16.}\newline
  \textbf{Reason:} \ldots{}\newline
  \textbf{Action:} \texttt{open coffeemachine 1}.\newline
  \textbf{Observation:} \texttt{Nothing happens.}\par\vspace{1pt}
  \textbf{Turns 17--30.}\newline
  \textbf{Reason on every turn:} \ldots{}\newline
  \textbf{Action on every turn:} \texttt{move mug 3 to coffeemachine 1}.\newline
  \textbf{Observation on every turn:} \texttt{Nothing happens.}\par
  \textbf{Outcome:} Turn limit reached.
}
\endgroup

\subsection{Using Observations to Choose the Next Action}
\label{app:reasoning-case-study}
The remaining cases compare how students use a failed-action observation (ALFWorld and ScienceWorld) or a product-page observation (WebShop) to choose the next action.

\subsubsection{ALFWorld: Recovering from a Failed Pickup}
\label{app:cd-failed-pickup}
The students share the first three actions, ending in a failed attempt to take a second CD. State-OPD changes action and succeeds in six turns; Vanilla OPD retries until the turn limit. Across four evaluations of this task, success is 3/4 versus 0/4.

\begingroup
    \footnotesize
    \tracepanel{traceTask}{Task and shared interaction}{%
      \textbf{Task.} \texttt{put two cd in drawer.}\par\vspace{1pt}
      \textbf{Obs 0:} ``-= Welcome to TextWorld, ALFRED! =-''\newline
      ``You are in the middle of a room. Looking quickly around you, you see a bed 2, a bed 1, a desk 1, a drawer 3, a drawer 2, a drawer 1, a garbagecan 1, a laundryhamper 1, and a shelf 1.''\newline
      ``Your task is to: put two cd in drawer.''\par\vspace{1pt}
      \textbf{Turn 1.}\newline
      \textbf{Reason:} \ldots{}\newline
      \textbf{Action:} \texttt{go to desk 1}.\newline
      \textbf{Observation:} ``You arrive at desk 1. On the desk 1, you see a alarmclock 1, a bowl 1, a cd 2, a cd 1, a cellphone 2, a cellphone 1, a mug 1, a pen 3, and a pencil 1.''\par\vspace{1pt}
      \textbf{Turn 2.}\newline
      \textbf{Reason:} \ldots{}\newline
      \textbf{Action:} \texttt{take cd 1 from desk 1}.\newline
      \textbf{Observation:} ``You pick up the cd 1 from the desk 1.''\par\vspace{1pt}
      \textbf{Turn 3.}\newline
      \textbf{Reason:} \ldots{}\newline
      \textbf{Action:} \texttt{take cd 2 from desk 1}.\newline
      \textbf{Observation:} \texttt{Nothing happens.}\par
    }
    \tracepanel{traceState}{State-OPD \hfill Success in 6 turns}{%
      \textbf{Turn 4.}\newline
      \textbf{Reason:} ``Wait, the correct command is "take (object) from (receptacle)". So the user tried "take cd 2 from desk 1", which is correct. But the system says nothing happens. Maybe the desk is not a receptacle? Wait, the desk is a receptacle. So perhaps the problem is that the user can only hold one item at a time. So after taking cd 1, the user can't take another item. So the solution is to put cd 1 into a drawer first, then take cd 2.''\par
      \textbf{Action:} \texttt{move cd 1 to drawer 1}.\newline
      \textbf{Observation:} ``You move the cd 1 to the drawer 1.''\par\vspace{1pt}
      \textbf{Turn 5.}\newline
      \textbf{Reason:} \ldots{}\newline
      \textbf{Action:} \texttt{take cd 2 from desk 1}.\newline
      \textbf{Observation:} ``You pick up the cd 2 from the desk 1.''\par\vspace{1pt}
      \textbf{Turn 6.}\newline
      \textbf{Reason:} \ldots{}\newline
      \textbf{Action:} \texttt{move cd 2 to drawer 1}.\newline
      \textbf{Observation:} ``You move the cd 2 to the drawer 1.''\par
      \textbf{Outcome:} Task completed.
    }
    \tracepanel{traceOpd}{Vanilla OPD \hfill Failure at the 30-turn limit}{%
      \textbf{Turn 4.}\newline
      \textbf{Reason:} ``Wait, the user's inventory is empty except for cd 1. So taking cd 2 should be possible. I'm stuck. Maybe the environment is not working correctly. But the user has to proceed. Let me try to take cd 2 again. Maybe the first attempt was a typo. Wait, the user wrote "take cd 2 from desk 1" correctly. So perhaps the environment is not allowing it.''\par
      \textbf{Action:} \texttt{take cd 2 from desk 1}.\newline
      \textbf{Observation:} \texttt{Nothing happens.}\par\vspace{1pt}
      \textbf{Turn 5.}\newline
      \textbf{Reason:} \ldots{}\newline
      \textbf{Action:} \texttt{take cd 2 from desk 1}.\newline
      \textbf{Observation:} \texttt{Nothing happens.}\par\vspace{1pt}
      \textbf{Turns 6--30.}\newline
      \textbf{Reason on every turn:} \ldots{}\newline
      \textbf{Action on every turn:} \texttt{take cd 2 from desk 1}.\newline
      \textbf{Observation on every turn:} \texttt{Nothing happens.}\par
      \textbf{Outcome:} Turn limit reached.
    }
\endgroup

\subsubsection{ScienceWorld: Recovering from Invalid Navigation}
\label{app:additional-case-studies}
\label{app:invalid-navigation}
Both students attempt an invalid move from the bathroom to the hallway. State-OPD returns via the kitchen and completes the task in ten turns; Vanilla OPD repeats the invalid move until turn 30.

\begingroup
\footnotesize
\tracepanel{traceTask}{Task and shared interaction through turn 3}{%
  \textbf{Task.} ``Your task is to find a(n) animal. First, focus on the thing. Then, move it to the blue box in the kitchen.''\par
  \textbf{Obs 0 (excerpt):} ``This room is called the kitchen. In it, you see: \ldots{} a blue box (containing nothing) \ldots{} You also see: A door to the bathroom (that is open) A door to the hallway (that is open) A door to the outside (that is open)''\par\vspace{1pt}
  \textbf{Turn 1.}\newline
  \textbf{Reason:} \ldots{}\newline
  \textbf{Action:} \texttt{go to bathroom}.\newline
  \textbf{Observation:} ``You move to the bathroom.''\par\vspace{1pt}
  \textbf{Turn 2.}\newline
  \textbf{Reason:} \ldots{}\newline
  \textbf{Action:} \texttt{look around}.\newline
  \textbf{Observation (relevant excerpt):} ``This room is called the bathroom. \ldots{} You also see: A door to the kitchen (that is open).''\par\vspace{1pt}
  \textbf{Turn 3.}\newline
  \textbf{Reason:} \ldots{}\newline
  \textbf{Action:} \texttt{go to hallway}.\newline
  \textbf{Observation:} \textbf{\texttt{No known action matches that input.}}\par
}
\tracepanel{traceState}{State-OPD \hfill Task completed in 10 turns}{%
  \textbf{Turn 4.}\newline
  \textbf{Reason:} \emph{(excerpt)} ``The user tried to go to the hallway but got an error message saying no known action matches that input. Hmm, maybe the hallway isn't directly accessible from the bathroom? Let me check the available commands again. \ldots{} The initial scene in the kitchen had doors to the bathroom, hallway, and outside. So, the hallway is adjacent to the kitchen, not the bathroom. So, from the bathroom, the user can't go to the hallway directly. The user needs to go back to the kitchen first, then go to the hallway.''\par
  \textbf{Action:} \texttt{go to kitchen}.\newline
  \textbf{Observation:} ``You move to the kitchen.''\par\vspace{1pt}
  \textbf{Turn 5.}\newline
  \textbf{Reason:} \ldots{}\newline
  \textbf{Action:} \texttt{go to outside}.\newline
  \textbf{Observation:} ``You move to the outside.''\par
  \textbf{Turn 6.}\newline
  \textbf{Reason:} \ldots{}\newline
  \textbf{Action:} \texttt{look around}.\newline
  \textbf{Observation (relevant excerpt):} ``Here you see: \ldots{} a blue jay egg \ldots{}''\par
  \textbf{Turn 7.}\newline
  \textbf{Reason:} \ldots{}\newline
  \textbf{Action:} \texttt{pick up blue jay egg}.\newline
  \textbf{Observation:} ``You move the blue jay to the inventory.''\par
  \textbf{Turn 8.}\newline
  \textbf{Reason:} \ldots{}\newline
  \textbf{Action:} \texttt{go to kitchen}.\newline
  \textbf{Observation:} ``You move to the kitchen.''\par
  \textbf{Turn 9.}\newline
  \textbf{Reason:} \ldots{}\newline
  \textbf{Action:} \texttt{focus on blue jay egg}.\newline
  \textbf{Observation:} ``You focus on the blue jay egg.''\par
  \textbf{Turn 10.}\newline
  \textbf{Reason:} \ldots{}\newline
  \textbf{Action:} \texttt{move blue jay egg to blue box}.\newline
  \textbf{Observation:} ``You move the blue jay to the blue box.''\newline
  \textbf{Outcome:} Task completed.\par
}
\tracepanel{traceOpd}{Vanilla OPD \hfill Failure at the 30-turn limit}{%
  \textbf{Turn 4.}\newline
  \textbf{Reason:} \emph{(excerpt)} ``Okay, the user tried to go to the hallway but got an error. Let me check the available commands. The available locations are the kitchen, bathroom, hallway, and outside. The user is in the bathroom, and the doors to the kitchen, hallway, and outside are open. The correct command should be "go to hallway" but maybe the system doesn't recognize that. Wait, the error message says "No known action matches that input." Maybe the correct command is "go to the hallway" with "the" included. Let me try that. \ldots{} The available commands include "go to LOC" where LOC is a location. So "go to hallway" should work. Let me try that again.''\par
  \textbf{Action:} \texttt{go to hallway}.\newline
  \textbf{Observation:} \texttt{No known action matches that input.}\par\vspace{1pt}
  \textbf{Turn 5.}\newline
  \textbf{Reason:} \ldots{}\newline
  \textbf{Action:} \texttt{go to hallway}.\newline
  \textbf{Observation:} \texttt{No known action matches that input.}\par\vspace{1pt}
  \textbf{Turns 6--30.}\newline
  \textbf{Reason on every turn:} \ldots{}\newline
  \textbf{Action on every turn:} \texttt{go to hallway}.\newline
  \textbf{Observation on every turn:} \texttt{No known action matches that input.}\par
  \textbf{Outcome:} Turn limit reached without completing the task.
}
\endgroup

\subsubsection{WebShop: Selecting Options Before Purchasing}
\label{app:webshop-option-selection}
Both students reach a product page that offers the requested red and size-8 options with neither selected. State-OPD selects both before purchasing, while Vanilla OPD buys immediately; across four evaluations of this task, success is 4/4 versus 0/4.

\begingroup
\footnotesize
\tracepanel{traceTask}{Task and common interaction through turn 3}{%
  \textbf{Instruction.} ``buy me some hiking boots with rubber soles. get them in red, size eight, and price lower than 60.00 dollars''\par\vspace{1pt}
  \textbf{Obs 0:} ``WebShop''\newline
  ``Instruction:''\newline
  ``buy me some hiking boots with rubber soles. get them in red, size eight, and price lower than 60.00 dollars''\newline
  ``[button] Search [button\_]''\par\vspace{1pt}
  \textbf{Turn 1.}\newline
  \textbf{Reason:} \ldots{}\newline
  \textbf{Action:} \texttt{search[hiking boots rubber soles red size 8 under 60 dollars]}.\par
  \textbf{Turn 2.}\newline
  \textbf{Reason:} \ldots{}\newline
  \textbf{Action:} \texttt{click[Next >]}.\newline
  \textbf{Observation:} Page 2 lists \texttt{B094H27MR3}, ``Women's Hiking Thermal Boots Outdoor Walking Shoes Non-Slip Hiking Shoes'', at ``\$21.99 to \$39.99''.\par
  \textbf{Turn 3.}\newline
  \textbf{Reason:} \ldots{}\newline
  \textbf{Action:} \texttt{click[B094H27MR3]}.\newline
  \textbf{Shared product-page observation:} the size buttons include \texttt{8}; the color buttons include \texttt{a | red} and \texttt{red}; \texttt{Buy Now} is available. No size or color button is yet marked as clicked.\par
}
\tracepanel{traceState}{State-OPD \hfill Requested variant purchased}{%
  \textbf{Turn 4.}\newline
  \textbf{Reason:} ``The color options include red, so I need to select that. The size options have 8 available. I should click on the red color and size 8. Then, check if the product meets all the criteria.''\par
  \textbf{Action:} \texttt{click[a | red]}.\newline
  \textbf{Observation:} ``You have clicked a | red.''\par\vspace{1pt}
  \textbf{Turn 5.}\newline
  \textbf{Reason:} ``Then I clicked on the red color option. Now, I need to select the size 8.''\par
  \textbf{Action:} \texttt{click[8]}.\newline
  \textbf{Observation:} ``You have clicked a | red. You have clicked 8.''\par\vspace{1pt}
  \textbf{Turn 6.}\newline
  \textbf{Reason:} \ldots{}\newline
  \textbf{Action:} \texttt{click[Buy Now]}.\newline
  \textbf{Outcome:} The requested color and size are purchased.\par
}
\tracepanel{traceOpd}{Vanilla OPD \hfill Required options omitted}{%
  \textbf{Turn 4.}\newline
  \textbf{Reason:} ``The next step is to click "Buy Now" to proceed with the purchase. Let me check if there are any other options, but the size and color are available. So, I should click "Buy Now" here.''\par
  \textbf{Action:} \texttt{click[Buy Now]}.\newline
  \textbf{Outcome:} The purchase does not include the requested color or size.\par
}
\endgroup

\end{document}